%% file: GeoMorph_arxiv.tex
\documentclass[journal]{IEEEtran}

\usepackage{amsmath,amssymb,amsfonts}
\usepackage{amsmath, amssymb, bm}

\usepackage{graphicx}
\usepackage{textcomp}

\usepackage{multirow}

\usepackage{caption}

\usepackage{subfigure}

\usepackage{hyperref}
\usepackage{hhline}

\usepackage{pifont}
\newcommand{\cmark}{\ding{51}}%
\newcommand{\xmark}{\ding{55}}%

\newcommand{\argmin}{\operatornamewithlimits{argmin}}

\begin{document}
\title{Unsupervised Multimodal Surface Registration with Geometric Deep Learning}
\author{Mohamed A. Suliman, Logan Z. J. Williams, Abdulah Fawaz, and Emma C. Robinson
\thanks{The authors are with the Department of Biomedical Engineering, School of Biomedical Engineering and Imaging Science, King’s College London, London, SE1 7EH, UK. Corresponding author: Mohamed A. Suliman (e-mail: mohamed.suliman@kcl.ac.uk).}
}

\include{macros}

\maketitle

\begin{abstract}
This paper introduces GeoMorph, a novel geometric deep-learning framework designed for image registration of cortical surfaces. The registration process consists of two main steps. First, independent feature extraction is performed on each input surface using graph convolutions, generating low-dimensional feature representations that capture important cortical surface characteristics. Subsequently, features are registered in a deep-discrete manner to optimize the overlap of common structures across surfaces by learning displacements of a set of control points. To ensure smooth and biologically plausible deformations, we implement regularization through a deep conditional random field implemented with a recurrent neural network. Experimental results demonstrate that GeoMorph surpasses existing deep-learning methods by achieving improved alignment with smoother deformations. Furthermore, GeoMorph exhibits competitive performance compared to classical frameworks. Such versatility and robustness suggest strong potential for various neuroscience applications. Code is made available at \href{https://github.com/mohamedasuliman/GeoMorph}{https://github.com/mohamedasuliman/GeoMorph}.
\end{abstract}


\section{Introduction}
\label{sec:introduction}
The human cortex is a highly convoluted and folded structure, exhibiting intricate variations in its topography among individuals \cite{amunts2000brodmann, glasser2016multi}. These variations pose significant challenges when aligning cortical surfaces for comparative analysis. 

Cortical surface registration aims to overcome these challenges by mapping data to a global average space, where common features of brain organization overlap across individuals. Such alignment facilitates various neuroimaging analyses including investigations of cortical morphometry, functional connectivity, neurodevelopment, and neurosurgical planning \cite{goubran2019multimodal, risholm2011multimodal, coalson2018impact}, to mention a few. Typically, image matching is driven towards alignment of univariate summary measures of cortical folding, such as sulcal depth or average surface curvature \cite{fischl2012freesurfer, yeo2009spherical, robinson2014msm}; however, in some cases, frameworks target alignment of cortical areas \cite{nenning2017diffeomorphic, robinson2014msm, robinson2018multimodal,abdollahi2014correspondences}. In recent years, numerous cortical surface registration algorithms and techniques have been developed, ranging from geometric-based methods to data-driven approaches, such as deep learning \cite{heinrich2020highly, dalca2019learning}. These methods employ various mathematical models, optimization algorithms, and similarity measures to align cortical surfaces accurately. 

Conventionally, cortical surfaces are mapped to a sphere as it captures the geodesic distances between points on the cortex. Registration is then performed by optimizing a similarity measure between the features on the target sphere and those on the deformed source sphere, while enforcing smoothness constraints. Freesurfer~\cite{fischl1999high} registers folding patterns on the surface by optimizing the mean-squared error (MSE) between a measure of average convexity across a set of subjects, and that of the individual, modulated by the inverse variance of the convexity across subjects. Spherical Demons (SD) \cite{yeo2009spherical} registers two spherical images in a coarse to fine manner by modifying the classical diffeomorphic Demons method \cite{vercauteren2009diffeomorphic}, initially implemented in the Euclidean image space, using velocity vectors tangent to the sphere. Multimodal Surface Matching (MSM) \cite{robinson2014msm,robinson2018multimodal} also implements matching in a coarse to fine manner, but chooses discrete optimization over classical approaches, as this conveys flexibility with regards to choice of cost function, and greater robustness to noise and local minima. In MSM, diffeomorphisms are encouraged by imposing a biomechanically-inspired, hyper-elastic strain regularisation on the deformation.
Combined these features were found to support multimodal (resting state fMRI and T1w/T2w ratio - myelin map) registration of cortical surfaces, in such a way that improved alignment of cortical areas when validated on independently collected task fMRI data \cite{coalson2018impact,glasser2016human,robinson2014msm,robinson2018multimodal,smith2013functional}. Note that all of the above methods solve an optimization problem for each pair of input images and, hence, exhibit long execution times. Also note that while diffeomorphisms have long been considered a pre-requisite for cortical surface registration, evidence has shown that cortical topography can vary in ways that break this assumption \cite{glasser2016multi,van2018parcellating}.
 
Recently, deep learning registration methods \cite{balakrishnan2019voxelmorph, dalca2019learning, de2019deep, heinrich2019closing, pielawski2020comir, shao2021prosregnet} have gained interest due to their faster execution times, improved ability to handle topographical variation, and efficiency in learning population-specific templates. These methods leverage the capabilities of deep neural networks to learn complex spatial transformations, and align cortical surfaces, with exceptional accuracy and efficiency.

Previous research has predominantly focused on learning-based registration frameworks for 2D or 3D Euclidean domains, such as brain volumes \cite{balakrishnan2019voxelmorph,dalca2019learning,de2019deep,fan2018adversarial}, lung CT \cite{heinrich2019closing,heinrich2020highly,fu2020lungregnet}, and histology \cite{borovec2020anhir,shao2021prosregnet,pielawski2020comir}. However, there is an increasing interest in adapting convolutional networks to non-Euclidean domains \cite{monti2017geometric,qi2017pointnet,zhao2019spherical}, leading to the development of learning-based registration methods for surfaces and point clouds \cite{aoki2019pointnetlk,wang2019deep,zhao2019spherical, cheng2020cortical}.

A noteworthy advancement in this field is the S3Reg framework \cite{zhao2021s3reg}, which learns displacements via the implementation of a spherical U-net network \cite{zhao2019spherical}, compares the overlap of moving and target features at baseline, and seeks to enforce diffeomorphisms using the scaling and squaring approach of the diffeomorphic Voxelmorph algorithm \cite{dalca2019learning}. A fundamental limitation of S3Reg is that the hexagonal filter implemented in \cite{zhao2019spherical} is not rotationally equivariant due to the lack of a global spherical coordinate system; hence, it flips directions at the poles and generates distortions. S3Reg overcomes this by using a combination of three networks, each trained on a different rotated version of the input. Recent studies have demonstrated that MoNet convolutions \cite{monti2017geometric}, learned from a mixture of Gaussian kernels, can achieve rotational equivariance \cite{fawaz2021benchmarking}. 

Motivated by S3Reg and inspired by deep-discrete registration frameworks \cite{heinrich2019closing,heinrich2020highly}, which handle large deformations, we propose a novel framework for spherical cortical registration, based on MoNet, named GeoMorph. We hypothesize that leveraging MoNet's rotational equivariance and its ability to learn larger deformations will enhance the generalization of our framework to brains with atypical topographies, and provide better registration results in terms of alignment quality. 

The process starts with feature extraction, which extracts low-dimensional feature representations for each input surface using MoNet graph convolutions. Inspired by MSM \cite{robinson2014msm,robinson2018multimodal}, the learned features are then registered in a deep-discrete manner by solving a multi-label classification problem, where each point in a low-resolution control grid deforms to one of a fixed, finite set of target locations, in such a way that maximises overlap between features across the two surfaces. To ensure smooth deformations, we impose a 
deep conditional random field (CRF), implemented using a recurrent neural network (RNN). This network updates the obtained deformation field in a way that ensures smoothness by forcing neighbouring points to deform similarly. 

Implementing GeoMorph in this way  presents distinct advantages with regards to its ability to register multimodal features simultaneously at the cortex. This paper extends our original deep discrete framework (DDR) \cite{suliman2022deep} through the introduction of a feature extraction network, which supports compact, low-dimensional representation of the input. This has proved fundamental to multimodal alignment, likely because it supports  weighting of different input features according to their significance. At the same time, the work in this paper expands from the preliminary analyses reported in \cite{suliman2022geomorph}, by validating the framework on multi-modal alignment of Human Connectome Project (HCP) and UK Biobank (UK) fMRI and T1w/T2w myelin features. Our objective is to demonstrate the versatility of the GeoMorph registration framework, showcasing its capacity to effectively align diverse cortical surface features and modalities. 
The results obtained from both univariate and multivariate experiments demonstrate that GeoMorph surpasses existing deep learning methods by achieving improved alignment accuracy and generating smoother and more biologically plausible deformations. Furthermore, GeoMorph exhibits competitive performance when compared to the best classical multimodal registration frameworks. 

\section{GeoMorph Architecture}
\label{sec: architecture}

\subsection{Background}
Let $\Mm, \Fm$ be the 3D coordinate matrices of the triangular meshes  of the moving ($M$) and fixed ($F$) images, formed on a sphere $\mathcal{S}^{2}$, centred on the origin; each has $N_{d}$ vertices, i.e., $\Mm, \Fm \in \mathbb{R}^{N_{d} \times 3}$. The objective of GeoMorph is to learn a spatial transformation $\Phim: M \to F$ that aligns the cortical features on $M$ to those on $F$ in the form
\begin{equation}
\Phim= \mathcal{F}_{\etav}\left(M, F\right)
\end{equation} 
upon optimizing a dissimilarity metric $\mathcal{L}$ 
\begin{equation}
\label{eq: regularization function}
\hat{\thetav} = \argmin_{\thetav} \mathcal{L} \left(\Phim_{\thetav};F,M\right) + \Sigma\left(\Phim_{\thetav}\right).
\end{equation} 
Here, $\mathcal{F}_{\etav}\left(\cdot\right)$ is a learnable function that is obtained using our geometric deep neural network GeoMorph, with $\etav$ being the network learnable parameters. The transformation $\Phim$ is parametrized with $\thetav$, while $\Sigma\left(\cdot\right)$ is a regularization function that imposes smoothness on $\Phim$. Finally, it is worth mentioning that we assume that data is presented to the network as concatenated cortical metric maps of $F$ and $M$, defined on a sphere $S^{2}$ that is parametrized by different resolutions (orders) of regularly sampled icospheres.

\subsection{Method Overview}

Let $\{\cv_{i}\}_{i=1}^{N_{c}} \in \mathbb{R}^{N_{c} \times 3}$ be the locations of $N_{c}$ control points on the moving sphere, generated from the vertices of a low-resolution icosphere $C \subset S^{2}$ (with $N_{c} << N_{d}$), and let $\{\lv_{i}\}_{i=1}^{N_{l}} \in \mathbb{R}^{N_{l} \times 3}$ represent the locations of $N_{l}$ label points, defined around each control point $\cv_{i}$, that represent all potential endpoints of the transformation $\mathcal{F_{\etav}}(\cv_{i})$. In all instances, the target labels are derived from the vertices of a higher-resolution icosphere (Fig.~\ref{fig: ico fig}a). The objective of GeoMorph is, therefore, to learn the optimal label (and hence displacement) for each control point to ensure features of the fixed and moving mesh are optimally aligned. Importantly unlike classical discrete frameworks \cite{robinson2014msm,robinson2018multimodal}, for which run-time is linked to the label dimensionality, GeoMorph is far less constrained by the extent of the label space. The general architecture of the GeoMorph network is shown in Fig.~\ref{fig: model}.

Learning on GeoMorph starts with feature extraction, which learns latent representations of features on $M$ and $F$ (Section~\ref{subsec: feature extraction net}). This is then followed by a classifier network which outputs $\Qm = \text{Softmax} \left(\Um\right) \in \mathbb{R}^{N_{c} \times N_{l}}$ softmax probabilities for each label around the control points (Section~\ref{subsec: classifier net}). Finally, the CRF-RNN network imposes smoothness on the learned deformation by encouraging neighboring control points to deform similarly (Section~\ref{subsec: crf-rnn net}).

\subsection{Network Preliminaries}

\subsubsection{Geometric convolutions}  As input features are assumed to lie on a spherical surface, we implement surface convolutions using Gaussian mixture models, as proposed in MoNet \cite{monti2017geometric}. Let $x$ be a vertex on the surface (or graph) and define $y \in \mathcal{N}\left(x\right)$ as a set of points in the neighbourhood of $x$, each associated with a multi-dimensional vector of pseudo-coordinates $\uv\left(x,y\right)$. Then, MoNet convolutions are defined as
\begin{equation}
\label{eq: conv 1}
\left(f \star g\right) \left(x\right) = \sum_{j} g_{j} D_{j}\left(x\right) f,
\end{equation}
where $f$ is the input feature map, $g$ is a learnable filter, and $D\left(x\right) f$ is a parametrisable patch operator given by
\begin{equation}
\label{eq: conv 2}
D_{j}\left(x\right) f = \sum_{y \in \mathcal{N}\left(x\right)}  w_{j} \left(\uv\left(x,y\right)\right) f\left(y\right), \forall j
\end{equation}
which extract the values of $f$ from the surface and then maps it at the neighborhood of $x$ using learnable filter weights $w_{j}$. The weights $w_{j}$ of MoNet are formulated using the a Gaussian function in the form
 \begin{equation}
\label{eq: conv 2}
w_{j}\left(\uv\right)= \exp\left(-\frac{1}{2} \left(\uv-\muv_{j}\right)^{T}\Sigmam_{j}^{-1} \left(\uv-\muv_{j}\right)\right),
\end{equation}
where $\Sigmam_{j} \in \mathbb{R}^{d \times d}$ and $\muv_{j} \in \mathbb{R}^{d \times 1}$ are a learnable covariance matrix and mean vector of a Gaussian kernel, respectively. 

\subsubsection{Icosphere resolutions} Starting from an icosphere of order $0$, which has 20 triangle faces, 30 edges, and 12 vertices, higher order resolution icospheres can be generated by hierarchically adding a new vertex to the center of each edge in each triangle (see Fig.~\ref{fig: ico fig}b). Let the number of the vertices at the current resolution level be $N$, then, the next higher resolution level will have $\left(N \times 4\right)-6$ vertices. In contrast, the previous lower resolution level will have $\left(N+6\right)/4$ vertices. For example, levels 1 and 2 have 42, and 162 vertices, respectively. Finally, note that with the exception of the first 12 vertices, which has only 5 neighbors, all vertices have 6 neighbors.

\subsubsection{Surface downsampling/upsampling} 
\begin{figure}[h]
\begin{center}
\includegraphics[width=2.5 in]{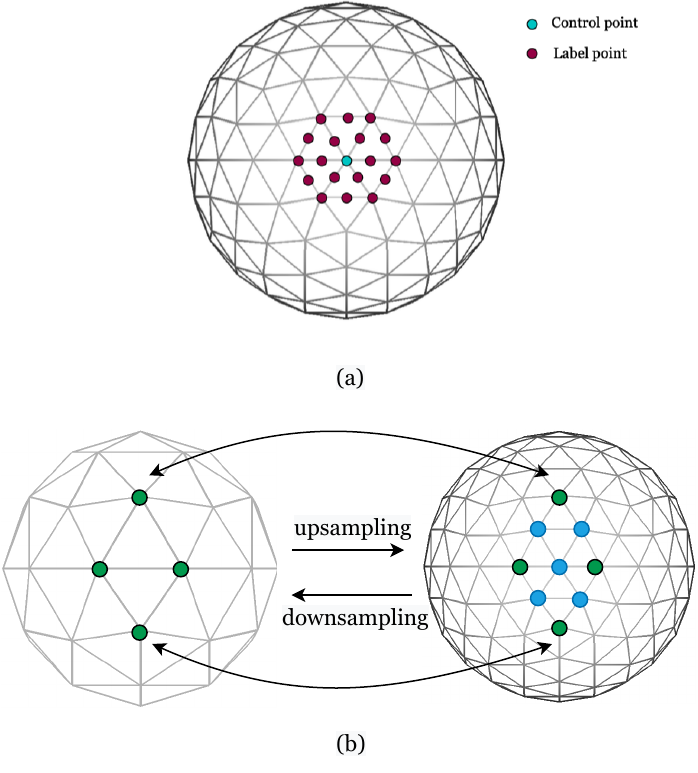}
\caption{a) Up and downsampling on icospheres. b) Example of a control point with its labels on the surface.} \label{fig: ico fig}
\end{center}
\end{figure}

\begin{figure*}[ht!]
\centering
\includegraphics[width=6. in]{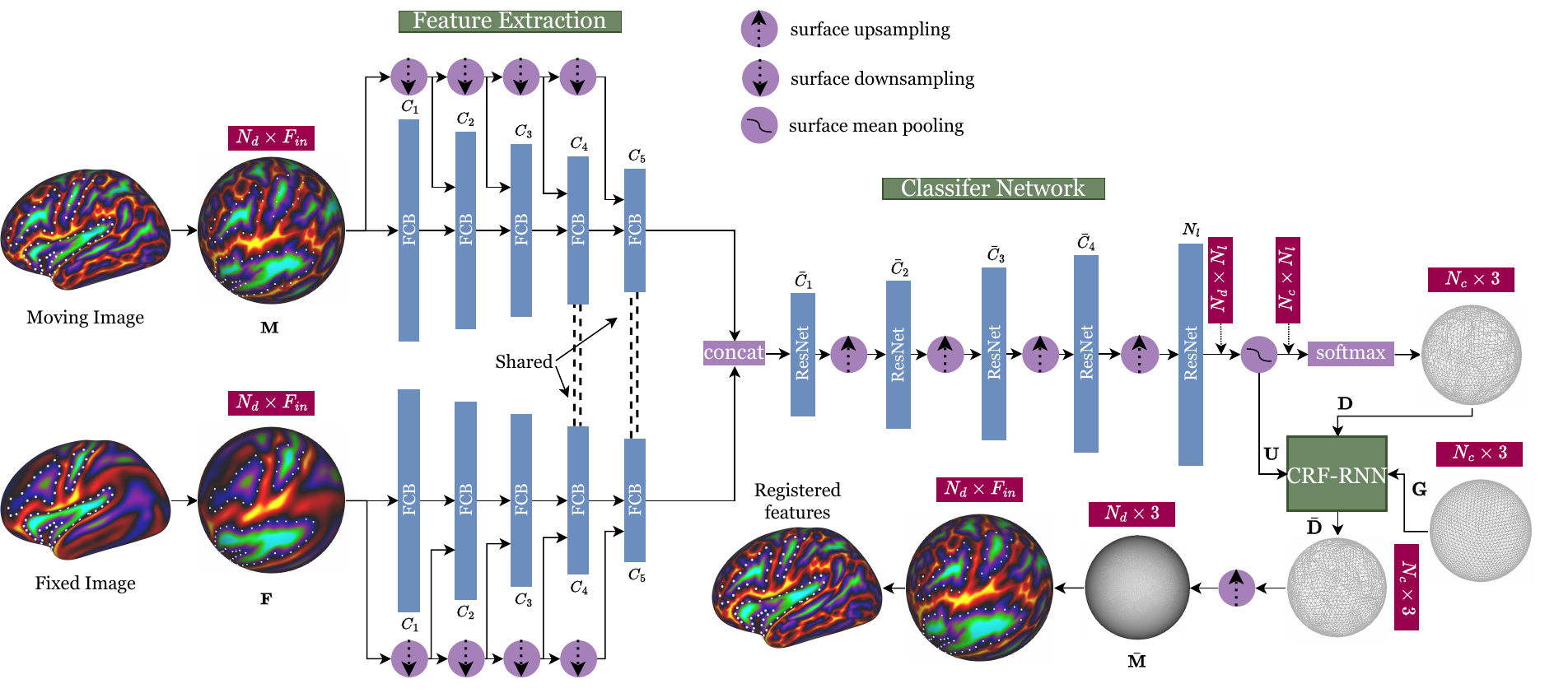}
\caption{GeoMorph network architecture. The dimensions in red boxes shows the input and the output dimensions at different network stages.} \label{fig: model}
\end{figure*}

Based on the icosphere nature above, we define downsampling as the process of extracting the vertices of the lower icosphere order from the current higher order icosphere. For the upsampling, i.e., $N \to \left(N \times 4\right)-6$, we obtain the next level icosphere by inserting new vertices as the average of their direct neighbours (see Fig.~\ref{fig: ico fig}b).

\subsubsection{Surface pooling} This is defined as the process of replacing each vertex and its neighbors by the mean or the max of the accumulated features from all of them. Hence, we obtain a downsampled icosphere $N \to \left(N+6\right)/4$ with new features.

\begin{figure*}[ht!]
\centering
\includegraphics[width=7. in, height=1.5 in]{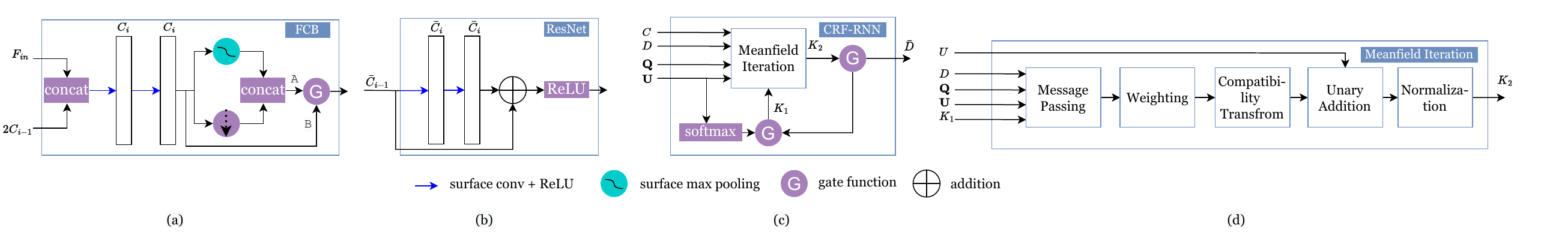}
\caption{a)  FCB architecture. b) ResNet architecture. c)  CRF-RNN architecture. d) Meafield Iterations architecture.} \label{fig: networks}
\end{figure*}


\subsection{Feature Extraction Network}
\label{subsec: feature extraction net}
The primary objective of this network is to learn low-dimensional feature representations for the features on $\Mm$ and $\Fm$, each on separate paths. The network takes the input features and the mesh topology in the form of vertex locations with neighborhood structure $\mathcal{N}$, where $\left(i,j\right) \in \mathcal{N} \subset \left\lbrace1, \dots, N_{d} \right\rbrace^{2}$ indicates that vertex $i$ is connected to $j$ by a triangle edge. A series of feature convolutional blocks (FCBs), each taking the previous stage output and a downsampled version of the input features, are applied to learn a low-dimensional feature space from each of $M$ and $F$, with only the weights of the last two FCBs being shared. The feature maps from each input are then concatenated to be passed to the classifier network.

At each FCB stage $i$ (Fig.~\ref{fig: networks}a), a total of $C_{i}$ features are learned, using a series of two MoNet convolutional filters along with spherical polar pseudo-coordinates, mean aggregation operators, and a LeakyReLU activation with parameter 0.2. The output features are then passed through a surface max pooling operator. To allow for global feature incorporation, the max pooling output is concatenated with a downsampled version of the LeakyReLU output. The result is then passed through a gate function $G$ with $G = A$ for $i=1, \dots,4$ and $G = B$ for $i=5$ (the last FCB block in the network) (see Fig.~\ref{fig: networks}a).

\subsection{Classifier Network}
\label{subsec: classifier net}
The learned features from the previous stage are passed through a series of five ResNet-inspired blocks, each learning $\bar{C}_{i}$ features, with the last one learning $N_{l}$ features. At each block, we perform two surface convolutions followed by a LeakyReLU activation with parameter 0.2 (see Fig.~\ref{fig: networks}b). The output of each network is first upsampled to the next icosphere order and then passed to the next stage. The output of the final ResNet, which is of dimension $N_{d} \times N_{l}$, is regularized through downsampling to the desired control grid resolution, i.e., $\Um \in \mathbf{R}^{N_{c} \times N_{l}}$. The optimal label assignment is then obtained from a softmax operation on $\Um$. Finally, we deform $\cv_{i}$ using the spherical coordinates of the labels to obtain a deformed control grid $D \subset S^{2}$.

\subsection{CRF-RNN Network} 
\label{subsec: crf-rnn net}
On its own, the classifier is of limited use since cortical registration is an ill-posed problem with many possible solutions. Moreover, the deformation from the classifier network does not incorporate any constraint; hence, it allows each control point to move independently, which could lead to deformations with high distortions, as we will show in Section~\ref{sec: results}.

Inspired by \cite{zheng2015conditional}, we introduce the CRF-RNN network in our architecture to impose smoothness by encouraging neighboring control points to deform to comparable label points. It takes as input: the control grid $C$, the deformed control grid $D$, the classifier network output $\Um$ and pseudo-probabilities $\Qm$, to output a regularized transformation grid $\bar{D}$. 

Let $Q_{\left(\cv_{i},\lv_{i}\right)}$ be the likelihood of deforming $\cv_{i}$ to the label point $\lv_{i}$. Moreover, we define the cost function $\varphi \left(\lv_{\cv_{i}},\lv_{\cv_{j}}\right); \varphi: S^{2} \to \mathbb{R},$ which measures the cost of deforming $\cv_{i}$ and $\cv_{j}$ to the label points $\lv_{i}$ and $\lv_{j}$, respectively. The CRF-RNN network optimizes the following CRF cost function
\begin{equation}
\label{equ: crf cost function}
E = \sum_{i} Q_{\left(\cv_{i},\lv_{i}\right)} +  \sum_{i \neq j}  \varphi \left(\lv_{\cv_{i}},\lv_{\cv_{j}}\right).
\end{equation}
Similarly to \cite{krahenbuhl2011efficient,zheng2015conditional}, we propose modelling $\varphi \left(\lv_{\cv_{i}},\lv_{\cv_{j}}\right)$ by
\begin{equation}
\label{eq: energy modelling}
\varphi \left(\lv_{\cv_{i}},\lv_{\cv_{j}}\right)= \mu \left(\lv_{i},\lv_{j}\right) K_{G}\left(\lv_{\cv_{i}},\lv_{\cv_{j}}\right),
\end{equation}
where $\mu$ is a learnable label compatibility function that captures correspondences between different pairs of label points, while $K_{G}$ is a Gaussian kernel \cite{krahenbuhl2011efficient,zheng2015conditional} of the form 
\begin{align}
\label{eq: gaussion kernel}
&K_{G}\left(\lv_{\cv_{i}},\lv_{\cv_{j}}\right) = \nonumber\\
&\omega\left(\cv_{i},\cv_{j}\right) \exp\left(-\frac{1}{2\gamma^{2}} \left(\lv_{\cv_{i}}-\lv_{\cv_{j}}\right)^{T} \Lambdam  \left(\lv_{\cv_{i}}-\lv_{\cv_{j}}\right) \right).
\end{align}
Here, $\omega$ are learnable filter weights, $\gamma$ is a kernel parameter, $\lv_{\cv_{i}}$ is the new spatial location of the deformed point $\cv_{i}$, while $\Lambdam$ is a symmetric, positive-definite, kernel characterization matrix. Note that our formulations in (\ref{eq: energy modelling}) and (\ref{eq: gaussion kernel}) are modified versions of those used in \cite{krahenbuhl2011efficient,zheng2015conditional}, with a single Gaussian kernel being used in this instance, and $\omega$ being introduced.

To minimize (\ref{equ: crf cost function}), we use the recurrent neural network (RNN) formulation of \cite{zheng2015conditional} (Fig.~\ref{fig: networks}c), which is based on multiple iterations of the mean-field CRF algorithm (Fig.~\ref{fig: networks}d).  Assuming that the CRF-RNN function is given by $\bar{f}_{\bar{\etav}}$, where $\bar{\etav}$ represents the network learnable parameters, and $T$ is the total number of mean-field iterations, the behavior of the network can be expressed by the following set of equations: 
\begin{equation}
\label{eq: cfr equation 1}
K_{1}\left(t\right) = \begin{cases}
  \text{Softmax}\left(\Um\right),\  t=0 \\      
  K_{2}\left(t-1\right), \ \ \ \ 0<t \leq T
\end{cases}
\end{equation}
\begin{equation}
\label{eq: cfr equation 2}
K_{2}\left(t\right) = \bar{f}_{\bar{\etav}}\left(\Um,C,D\right)
\end{equation}

\begin{equation}
\label{eq: cfr equation 3}
Y\left(t\right)= \begin{cases}
  0, \ \ \  \  \ \ \ \ \ 0\leq t < T\\      
  K_{2}\left(t\right), \ \ t = T
\end{cases}
\end{equation}
As Fig.~\ref{fig: networks}d shows, the mean-field algorithm starts at the message passing stage. This applies filter $K_{G}$ to $\Qm$. The weights of $K_{G}$ are learned on the weighting stage, which can be viewed as a convolution with a $1 \times 1$ filter that has $N_{c}$ input and output channels (i.e., learning $\omega$). The outputs from this stage are then shared between the labels, depending on the compatibility between them, with all pairs being assigned a different penalty (i.e., learning $\mu$). This operation can be performed by using a convolutional layer with a filter that has $1 \times 1$ receptive field and a total of $N_{l}$ input and output channels. Next, we update $\Um$ using the output from the compatibility stage by subtracting its values from $\Um$. Finally, a normalization operation, using a softmax function, is performed. In all these stages, we can easily show that we can calculate error differentials with respect to the input. Thus, we can train the CRF-RNN network end-to-end utilizing the back-propagation algorithm. It is also worth mentioning that it is shown in \cite{krahenbuhl2011efficient,zheng2015conditional} that the mean-field iterative algorithm converges in less than 10 iterations.  Once complete, the regularized deformed control grid $\bar{D}$ from the CRF-RNN network is upsampled to the input resolution using barycentric interpolation, to obtain $\bar{M} \subset S^{2}$. Features from the moving image are then resampled from $\bar{M}$ to $F$ using adaptive barycentric interpolation \cite{glasser2013minimal,robinson2018multimodal}, implemented through Workbench Command \cite{marcus2011informatics}.

\subsection{Loss Function}
The network optimization is derived using an unsupervised loss function $\mathcal{L}$ of the form:
\begin{equation}
\label{eq: loss func}
\mathcal{L} \left(\Phim;F,M\right) = \lambda_{\text{sim}} \mathcal{L}_{\text{sim}} \left(F,M\right) + \lambda_{\text{sm}} \mathcal{L}_{\text{sm}}\left(\Phim\right) .
\end{equation}
Here, $\mathcal{L}_{\text{sim}}$ measures the similarity between the features on $F$ and those on $\bar{M}$. We use a measure that is a sum of the MSE and cross-correlation (CC), i.e., 
\begin{equation}
\mathcal{L}_{\text{sim}}  = \frac{1}{N_{d}}\sum_{i=1}^{N_{d}} \left( \left| \left| F_{\vv_{i}} - \bar{M}_{\vv_{i}}\right|\right|_{2}^{2} - \frac{cov\left(F_{\vv_{i}}, \bar{M}_{\vv_{i}}\right) }{\sigma_{F_{\vv_{i}}} \sigma_{\bar{M}_{\vv_{i}}}} \right),
\end{equation}
where $F_{\vv_{i}}, \bar{M}_{\vv_{i}}$ denote the corresponding features at vertex $i$, $cov\left(\cdot,\cdot\right)$ is the covariance operator, while $\sigma$ is the standard deviation measure. On the other hand, the term $\mathcal{L}_{\text{sm}}$ is introduced to allow for more user control over the balance between accurate alignment and smooth deformation and is formulated as a diffusion regularization penalty on the gradients of the $\Phim$, i.e., $\mathcal{L}_{\text{sm}} =\left( \left|\bigtriangledown \Phim_{\xv} \right| + \left| \bigtriangledown\Phim_{\yv} \right| + \left|\bigtriangledown \Phim_{\zv} \right| \right)$, where $\xv, \yv, \zv$ refer to the cardinal directions, while $\bigtriangledown$ is the gradient operator. Hence, $\mathcal{L}_{\text{sm}} = \Sigma$ from (\ref{eq: regularization function}). To compute $\bigtriangledown$, we apply the hexagonal filter from \cite{zhao2019spherical}, which approximates spherical gradients on spherical surfaces (see \cite[Fig. 4]{zhao2019spherical} for more information). Finally, $ \lambda_{\text{sim}} \geq 0$ and $\lambda_{\text{sm}} \geq 0$ are hyperparameters.


\section{Experiments}
\label{sec: experiments}
To validate GeoMorph, we conducted a series of experiments on real data collected as part of the adult Human Connectome Project (HCP) \cite{glasser2013minimal} and the UK Biobank (UKB) \cite{miller2016multimodal,alfaro2018image}. In each case, $M$ represents cortical features from an individual subject, whereas $F$ represents features from a fixed population average atlas. We validate GeoMorph for both univariate alignment of cortical folding (sulcal depth) features, and multivariate alignment of T1w/T2w myelin maps and coarse scale surface RSNs (equivalent to MSMAll). Only left hemisphere surfaces were used.

\subsection{Datasets}
\subsubsection{HCP}
The HCP dataset consists of cortical feature maps and meshes, derived from 1110 individuals, aged between 22 and 35 years. Participants were scanned over a two-day visit at Washington University in St. Louis, with a customized 3-Tesla Siemens Skyra, using a 32-channel head coil. Data used in this study included features derived from structural MRI and both task and resting-state functional MRI. 
Structural MPRAGE (T1w) and SPACE (T2w) scans were acquired with 0.7 mm isotropic acquisitions, whereas all fMRI was obtained with a resolution of 2 mm isotropic. In total, four resting state fMRI (rfMRI) scan sessions were acquired (15 minute of acquisition per session), with a repetition time (TR) of 0.720s, multiband factor 8, resulting in 4800 timepoints (1200 timepoints per session per subject). A total of seven tasks were performed: emotional, gambling, language, motor, relational, social cognition, and working memory\footnote{More details are available at \cite{barch2013function, van2013wu}.}. The tfMRI were acquired after the rfMRI scans. Hence, these datasets represent entirely independent sets, allowing tfMRI to be be used to robustly validate the performance of multimodal alignment for cortical areas, as demonstrated in numerous prior studies, e.g., \cite{coalson2018impact, glasser2016multi, robinson2018multimodal}.

\subsubsection{UKB}
The UKB datasets consists of comparable features derived from 3000 UKB subjects, aged between 46 and 83 years. Scans were acquired using a 3-Tesla Siemens Skyra scanner and Siemens 32-channel head coil at 4 different locations in the UK. Structural images of 1 mm isotropic T1w and T2-FLAIR were acquired, whereas functional images were obtained at a resolution of 2.4 mm isotropic and for a total duration of 6 minutes. A single resting-state fMRI session was acquired, with a TR of 0.735s, multiband factor 8, resulting in 490 timepoints per subject \cite{miller2016multimodal,alfaro2018image}. 

\subsection{Preprocessing}
\label{sec: preprocessing}
In all cases, cortical surfaces were reconstructed using FreeSurfer \cite{fischl2012freesurfer} following the HCP Structural Pipelines \cite{glasser2013minimal,williams2023geneneralising}. Cortical surfaces were extracted using both T1w and T2w images, which improves placement of the pial surface \cite{glasser2013minimal}. T1w/T2w maps were generated using the volumetric bias correction method described by \cite{glasser2011mapping}, with additional removal of low frequency biases in T1w/T2w across the cortical surface \cite{glasser2013minimal}. Resting-state fMRI was motion and distortion corrected, high-pass filtered, intensity normalised, and registered to MNI template space using FNIRT \cite{glasser2013minimal,alfaro2018image}. Structured noise was removed from rfMRI timeseries using ICA-FIX \cite{salimi2014automatic,griffanti2014ica}. Cleaned timeseries were mapped to the cortical surface with ribbon-constrained volume-to-surface mapping \cite{glasser2013minimal}. 

Task-fMRI preprocessing was the same as rfMRI. Single subject task analysis was first modelled within-run (first-level analysis), then between-runs (second-level analysis) using fixed effects general linear model with FSL FEAT \cite{woolrich2001temporal}. Group level task analyses were then performed using a mixed effects general linear model \cite{woolrich2004multilevel}. Outputs were then projected to the cortical surface using ribbon-constrained volume-to-surface mapping, and minimally smoothed on the cortical surface, using a kernel of 2mm FWHM \cite{glasser2013minimal}. 

Given HCP and UKB data were acquired using different techniques, we first matched the histograms of the cortical features of the UKB subjects to those of the HCP subjects. Then, all features were normalized within-subject to a zero mean and a standard deviation of one, with their extreme values being clipped at ± 2 standard deviations of their respective distributions. The medial wall of the cortical surface, which does not contain any cortical grey matter and represents a combination of cerebrospinal fluid, white matter and non-cortical grey matter, was considered an artifact and was masked out.

Experiments validating multimodal GeoMorph were compared against MSMAll \cite{robinson2018multimodal,glasser2016human}. For this cortical surface data were first coarsely aligned based on cortical folding (MSMSulc), then alignment was driven using a combination of 32 RSN spatial maps and T1w/T2w myelin. RSNs were derived from weighted dual regression of group ICA spatial maps (dimension = 40) \cite{glasser2016multi}. Cortical features were then resampled to a regular icosphere of order six (with 40,962 equally spaced vertices) using barycentric interpolation. Both myelin and functional data in HCP and UKB were smoothed on the surface using a 4 mm FWHM geodesic Gaussian smoothing kernel.

\subsection{Implementation}
\label{sub: geomorph Implementation}
GeoMorph was implemented in PyTorch, with MoNet convolutions derived from the PyTorch Geometric library \cite{Fey/Lenssen/2019}, and the number of kernels being set to 10. 
In all experiments, optimisation was performed using ADAM \cite{kingma2014adam} and the mean-field iterative algorithm in the CRF-RNN network was set to 5 iterations. Network configurations were different for unimodal and multimodal registration as follows:

\subsubsection{Univariate registration} Registration was driven using sulcal depth as a feature and was optimised, in a coarse-to-fine fashion, by two GeoMorph networks that were trained serially; the first of which optimises alignment for a low resolution control point grid derived from an icosphere of order 2 (with $N_{c} = 162$), with $N_{l} = 600$ labels generated from an icosphere of order 5; the second refines alignment for a higher resolution control point grid, corresponding to  an icosphere of order 4 ($N_{c} = 2542$), with $N_{l} = 1000$ label vertices that were generated from an icosphere of order 8. A total of 1110 cortical surfaces with sulcal depth features from the HCP were used in this experiment. A split of 888-111-111 train-validation-test was implemented in all experiments with batch size being set to 1.
Network parameters for the first network were set to: $C_{1}= 32, \gamma = 0.7, \lambda=1.5, r=1,$ and $\bar{C}_{1}=600$; whereas for the second network: $C_{1}= 2, \gamma = 0.2, \lambda=0.6, r=5$, and we set $[\bar{C}_{i}]_{i=1}^{5} = [8,16,64,128,1000]$. The learning rate was set to $10^{-3}$. The coarse network was trained for 100 epochs, each time learning the deformation at the control grid level, upsampling it to the input resolution level using barycentric interpolation, and then resampling the input sulc features to this newly deformed sphere to compare with the fixed image sulc features. Once training was done, the network parameters that provided the best validation score were saved. The resulted deformed moving image was then passed to the fine stage, and this new network was trained for 100 epochs. The final performance on the test set was reported using network parameters that provided the best validation score. 

\subsubsection{Multimodal registration} Training was performed using myelin and RSNs derived from both HCP and UKB, with the learning rate being set to $2e^{-4}$. The batch size was set to 1 for all experiments and train-validation-test splits of 801-100-100  and 2556-200-200 were used for the HCP and UKB respectively. The HCP and UKB data were stacked together and then randomly shuffled during the training and the validation phase.  Unlike univariate registration, no further improvement in performance was observed for multi-stage image registration; hence, a single network with high resolution control point grid was used, set to the resolution of an order 4 icosphere i.e., $N_{c} = 2542$, with $N_{l} = 600$ lying on an icosphere of order $6$. Moreover, we let $[C_{i} ]_{i=1}^{5} = [32, 32, 64, 64, 128], \gamma = 0.2, \lambda=0.6, r=5$, and we set $[\bar{C}_{i}]_{i=1}^{5} = [256,128,128,128,600]$. Training was carried out in two stages. First, the network was pretrained using an autoencoder whose architecture mirrored that of the feature extraction network - with an identical encoder and equivalent decoder layers implemented in reverse. Following that, the GeoMorph network was trained for 100 epochs and the network performance with the best validation score was reported. A discussion on the impact of the various parameter sets mentioned above can be found in Section~\ref{sec:ablation}.

\subsection{Benchmark Methods}
GeoMorph was benchmarked against SD, MSM, Freesurfer, and the learning-based method S3Reg. The validation was performed using the official implementations of SD\footnote{\href{https://github.com/ThomasYeoLab/CBIG}{https://github.com/ThomasYeoLab/CBIG}}), MSM Pair\footnote{Available through FSLv6.0}, MSM Strain\footnote{\href{https://github.com/ecr05/MSM_HOCR}{https://github.com/ecr05/MSM$\_$HOCR}}, and S3Reg\footnote{\href{https://github.com/zhaofenqiang/SphericalUNetPackage}{https://github.com/zhaofenqiang/SphericalUNetPackage}}.

\subsubsection{Univariate registration} 
To achieve fair comparison across all methods, the hyper-parameters were tuned for each, and performance across all parameter configurations were reported. The following parameters were optimised:
\begin{itemize}
    \item SD: The number of smoothing iterations used to smooth the final displacement field (in the Spherical Demons second step) was selected from  $\left[1,5,10\right]$, whereas the smoothing variance $\sigma_{x} $ was varied over $\left[1,2,6,10\right]$ (hence, ending with a total of 11 experiments).
    \item MSM: Two versions of MSM were used: MSM Pair \cite{robinson2014msm}, which uses first-order (pairwise) penalties, and MSM Strain \cite{robinson2018multimodal}, which applies high-order smoothness constraints derived from physically relevant equations of strain energy. A single regularization parameter was set at the 4 stages of the registration (coarse to fine). Here the regularization was varied over 22 weighting factors that differed across MSM Pair ($\lambda \in \left[0.0001,  0.2\right]$) and MSM Strain configurations ($\lambda \in \left[0.0001,  0.9\right]$).
    \item S3Reg: The network employs distinct regularization parameters at each of its 4 registration stages. To achieve optimal performance, 7 experiments were conducted, each utilizing specific configurations of regularization penalties for the corresponding sets: (refer to \cite{zhao2021s3reg} for more information): $[2,5,6,8]$, $[2,10,12,20]$, $[2,10,12,14]$, $[2,5, 12,16]$, $[2,10,6,8]$, $[2,10,12,8]$, and $[2,5,6,16]$. In each case, S3Reg networks were trained for 100 epochs at each registration level, and the performance of the network with the best validation score was reported.
    \item Freesurfer: The method is not tunable, and therefore results were reported for its default parameterization.
\end{itemize}
Note that all these frameworks perform coarse-to-fine, multi-stage registration over 4 icosphere resolutions. Moreover, S3Reg framework has an additional spherical transform network that seeks to enforce a diffeomorphic registration.


\subsubsection{Multimodal registration}  
Multimodal experiments benchmark GeoMorph solely against MSM - as the most highly optimised and rigorously benchmarked classical framework for multimodal image registration. In this comparison, two variants of MSM were used: namely MSMSulc and MSMAll, as outlined in Section~\ref{sec: preprocessing}. Moreover, two versions of GeoMorph were presented: GeoMorphSulc which denotes alignments achieved using GeoMorph driven by sulcal depth features, and GeoMorphAll which corresponds to alignments obtained utilizing myelin and rfMRI features from both HCP and the UKB datasets. The inclusion of GeoMorphSulc aims to showcase the improvements attained through multimodal registration.

\subsection{Evaluation Measures}
The performance of all methods was compared based on their alignment quality, assessing how well features in the source and target meshes overlap using cross-correlation (CC) similarity, and also on the smoothness of the resulted deformation using areal and shape distortions. The distortions are calculated from the local deformation ($\Fm_{pqr}$) of each triangular face, defined by vertices $\pv, \qv, \rv$. The eigenvalues of $\Fm$ ($\lambda_1$ and $\lambda_2$) represent principal in-plane stretches \cite{knutsen2010new}, such that relative change in the area may be described by $J=\lambda_1/\lambda_2$, whereas the relative change in shape may be described by $R=\lambda_1/\lambda_2$. The areal distortion is defined as $\log_{2}J = \log_{2} \left(\text{Area}_1/\text{Area}_2\right)$, while shape distortion is defined by $\log_{2}R$. Each of these measures was measured across all registered surfaces. 

Multimodal registration derived with myelin and rfMRI was also evaluated using HCP tfMRI data. In this case, improvements in alignment were assessed qualitatively and quantitatively upon comparing the group mean activation maps using a `cluster mass' measure \cite{robinson2014msm,glasser2016multi}; this quantifies the size of the supra-threshold clusters and the magnitude of the statistical values within them, and is obtained using the following formula: $CM = \sum_{i \in \mathcal{T}} \| z\left(\xv_{i}\right) \| A\left(\xv_{i}\right)$. Here, $\xv_{i}$ is the vertex coordinate, $z\left(\xv_{i}\right)$ is the statistical value at $\xv_{i}$, $A\left(\xv_{i}\right)$ is one third of the area associated with $\xv_{i}$ \cite{winkler2012measuring}, and $\mathcal{T}$ represents the set of vertices with $| z\left(\xv_{i}\right) | \geq 5$. The area $A\left(\xv_{i}\right)$ is obtained from a share of the area of each mesh triangle connected to it in the mid-thickness surface. The cluster mass was obtained for each contrast within the set of 7 HCP task experiments, a total of 86 contrasts. Note that a higher cluster mass measure indicates a better registration performance.  

\begin{table*}[!ht]
\begin{center}
\caption{Distortions measures and average runtime for different methods at CC $\sim 0.88$. Classical methods (top) and learning-based methods (bottom).}\label{table: dist}
\centering
    \begin{tabular}{cc||c|c|c|c||c|c|c|c||c|c|}
        \hline
        \hline
        \multicolumn{1}{|c|}{\multirow{2}{*}{\textbf{Methods}}}          & \multirow{2}{*}{\textbf{\begin{tabular}[c]{@{}c@{}}\textbf{CC}\\ \textbf{Similarity}\end{tabular}}} & \multicolumn{4}{|c||}{\textbf{Areal Distortion}}  & \multicolumn{4}{c||}{\textbf{Shape Distortion}} & \multicolumn{2}{c|}{\textbf{Avg. Time}}  \\ \cline{3-12}
        \multicolumn{1}{|c|}{}            &   &\textbf{Mean}   & \textbf{Max}       & \textbf{95\%} & \textbf{98\%}                           &\textbf{Mean }      &\textbf{Max}       & \textbf{95\%}  & \textbf{98\%}  & \textbf{CPU}         &\textbf{GPU}  \\ \hline
        \hline
        \multicolumn{1}{|c|}{\begin{tabular}[c]{@{}c@{}}  Freesurfer  \end{tabular}}   
        
              &  0.75   & 0.34  & 11.73 & 0.82     &    1.00 & 0.63   &6.77  &1.29   &1.54   & 30 min   & -   \\                            
                    
        \multicolumn{1}{|c|}{\begin{tabular}[c]{@{}c@{}} MSM Pair \end{tabular}}
        
         & 0.877        &0.41   & 9.17 & 1.24      &1.76  &0.62   & 9.05  &1.61   &2.16   & 13 min    & -     \\ 
     
     \multicolumn{1}{|c|}{\begin{tabular}[c]{@{}c@{}} MSM Strain \end{tabular}}
        
         & 0.880      & 0.27   &\textbf{1.06}  & 0.53       & 0.66    &0.64    &\textbf{1.93}  &1.17  &1.30   &  1 hour   & -        \\ 
         
         \multicolumn{1}{|c|}{\begin{tabular}[c]{@{}c@{}}  SD  \end{tabular}}   
        
        & 0.875 & \textbf{0.18} &  2.00  &\textbf{0.50}   &\textbf{0.65} &\textbf{0.24}  &1.98  &\textbf{0.50}  &\textbf{0.65}   & \textbf{1 min}  &  -    \\ \hline \hline
               
         \multicolumn{1}{|c|}{\begin{tabular}[c]{@{}c@{}}  S3Reg \end{tabular}}
        
         & 0.875      & 0.26   &22.22 & 0.82   & 1.16  &0.51    &  21.65  &1.35   &2.0    & 8.8 sec         &    8.0 sec    \\ 
         
         \multicolumn{1}{|c|}{\begin{tabular}[c]{@{}c@{}} GeoMorph \end{tabular}}
        
         & 0.875    & \textbf{0.19}  &\textbf{2.43}   &\textbf{0.53}  & \textbf{0.69}  &\textbf{0.26}  &\textbf{2.70}    &\textbf{0.63}   &\textbf{0.82}    & \textbf{8.3 sec} &   \textbf{2.6 sec }  \\ \hline \hline
    \end{tabular} 
    \end{center}  
\end{table*}

\section{Results}
\label{sec: results}
\subsection{Univariate Registration}

\begin{figure}[h]
\begin{center}
\includegraphics[width=3 in]{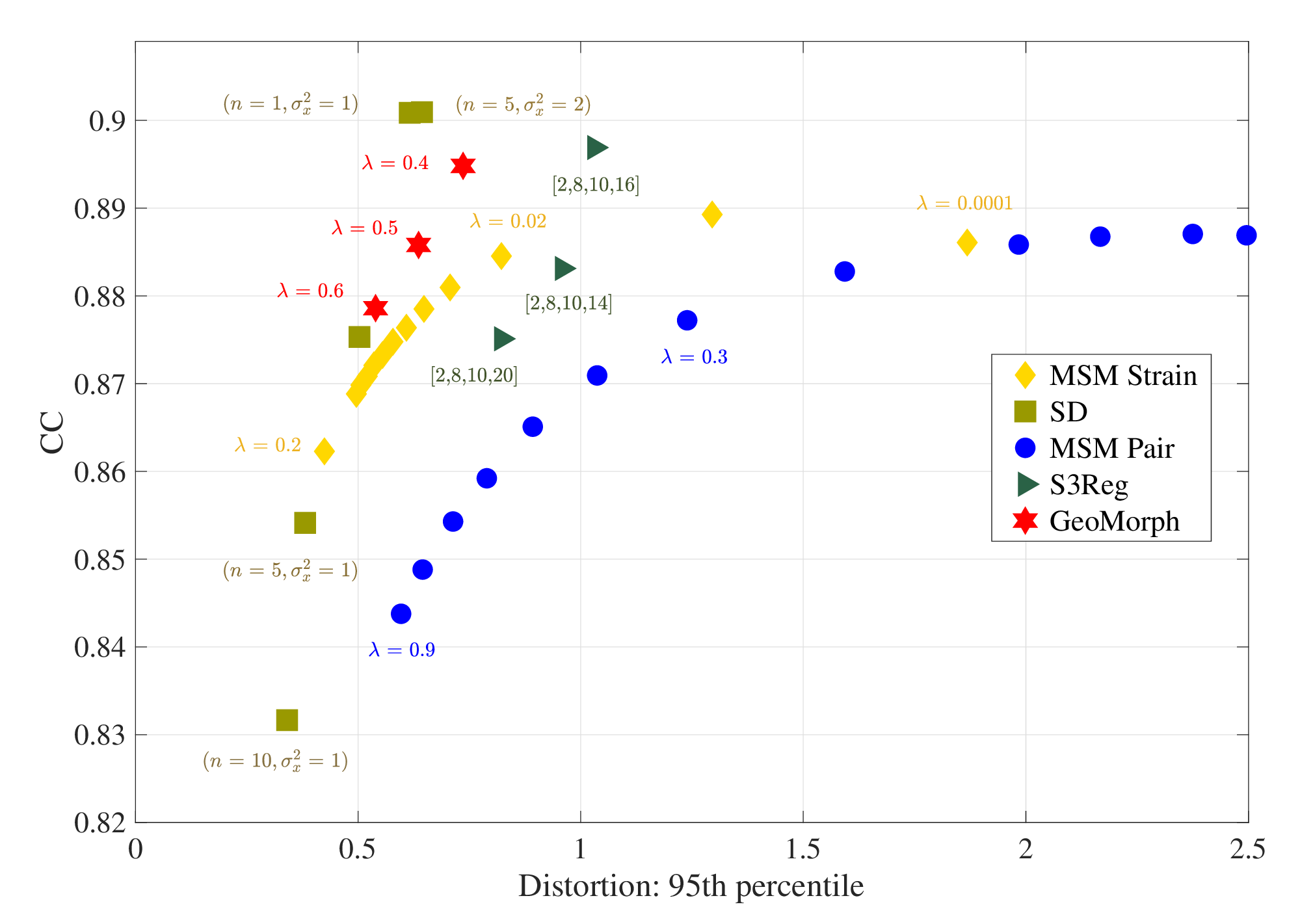}
\caption{Similarity performances of all methods vs. the 95th percentile of the areal distortion at multiple regularization levels across runs.} \label{fig: methods per vs runs}
\end{center}
\end{figure}

Fig.~\ref{fig: methods per vs runs} illustrates the similarity performances of various runs of all methods plotted against the 95th percentile of the absolute value of the areal distortion. For each similarity level, GeoMorph exhibits distortions falling within the range of the best classical methods (SD and MSM Strain) and demonstrates reduced extremes of areal distortions compared to S3Reg. Table.~\ref{table: dist} summarises the performance of all surface registration frameworks on the task of sulcal depth alignment. In each case, results are reported for the configuration that generated a mean CC value of approximately 0.88 (as this is the best CC value that all methods can achieve). Performance should therefore be judged in terms of which methods achieve the lowest mean, maximum, 95th percentile, and 98th percentile values of the distortion. In this, GeoMorph is the second best performing framework (behind SD but better than MSMStrain). On the other hand, S3Reg and MSMPair exhibit much poorer performance. It is worth noting that these values represent the optimal performance of S3Reg, across all runs. In terms of average run time, when utilizing a PC equipped with an NVIDIA Titan RTX 24GB GPU and an Intel Core i9-9820X 3.30 GHz CPU, GeoMorph exhibits the least GPU and CPU times among all methods.

\begin{figure}[h!]
  {%
    \subfigure[Areal Distortion.]{\label{fig:strian_j}%
      \includegraphics[width=1\linewidth, height =2.3in]{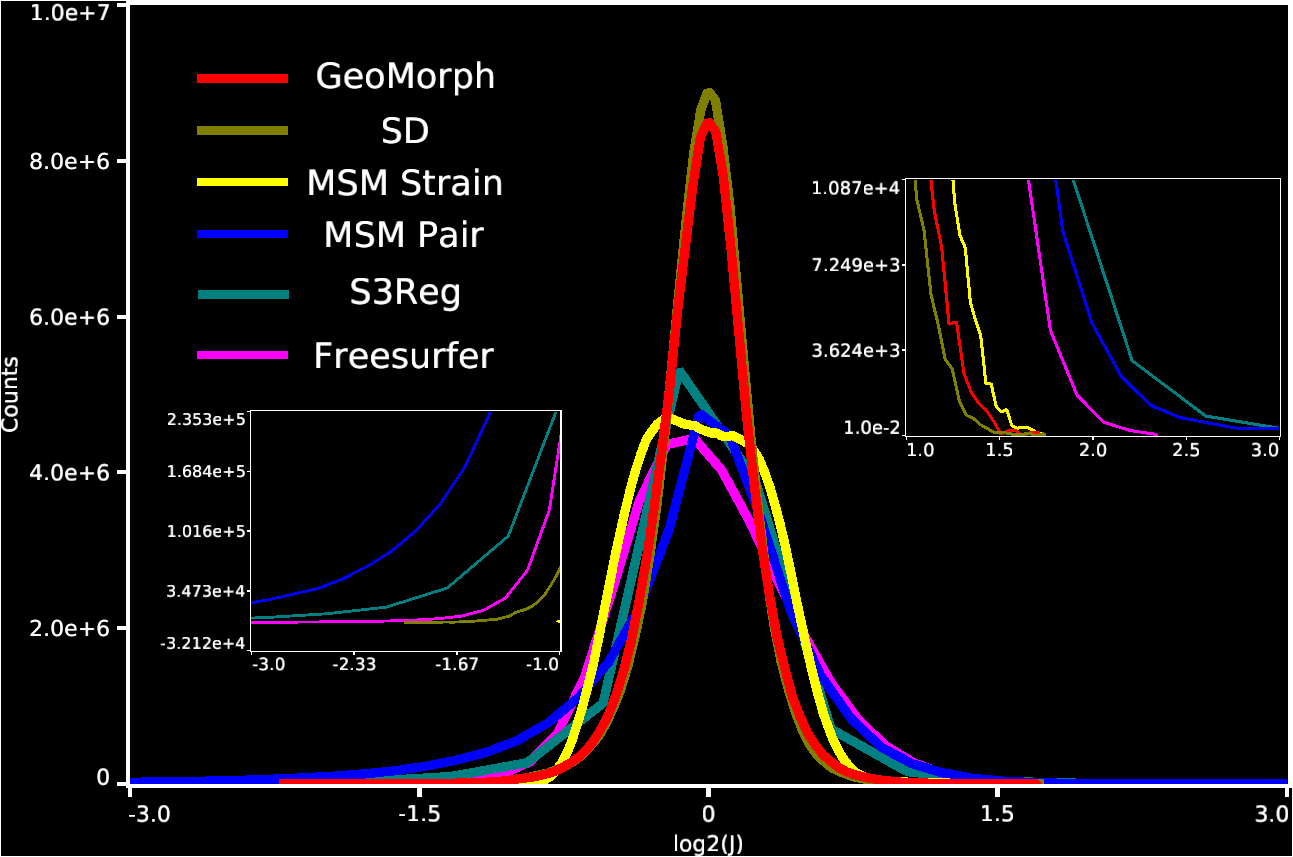}}%
    \qquad
    \subfigure[Shape Distortion.]{\label{fig:strain_r}%
      \includegraphics[width=1.\linewidth, height =2.3in]{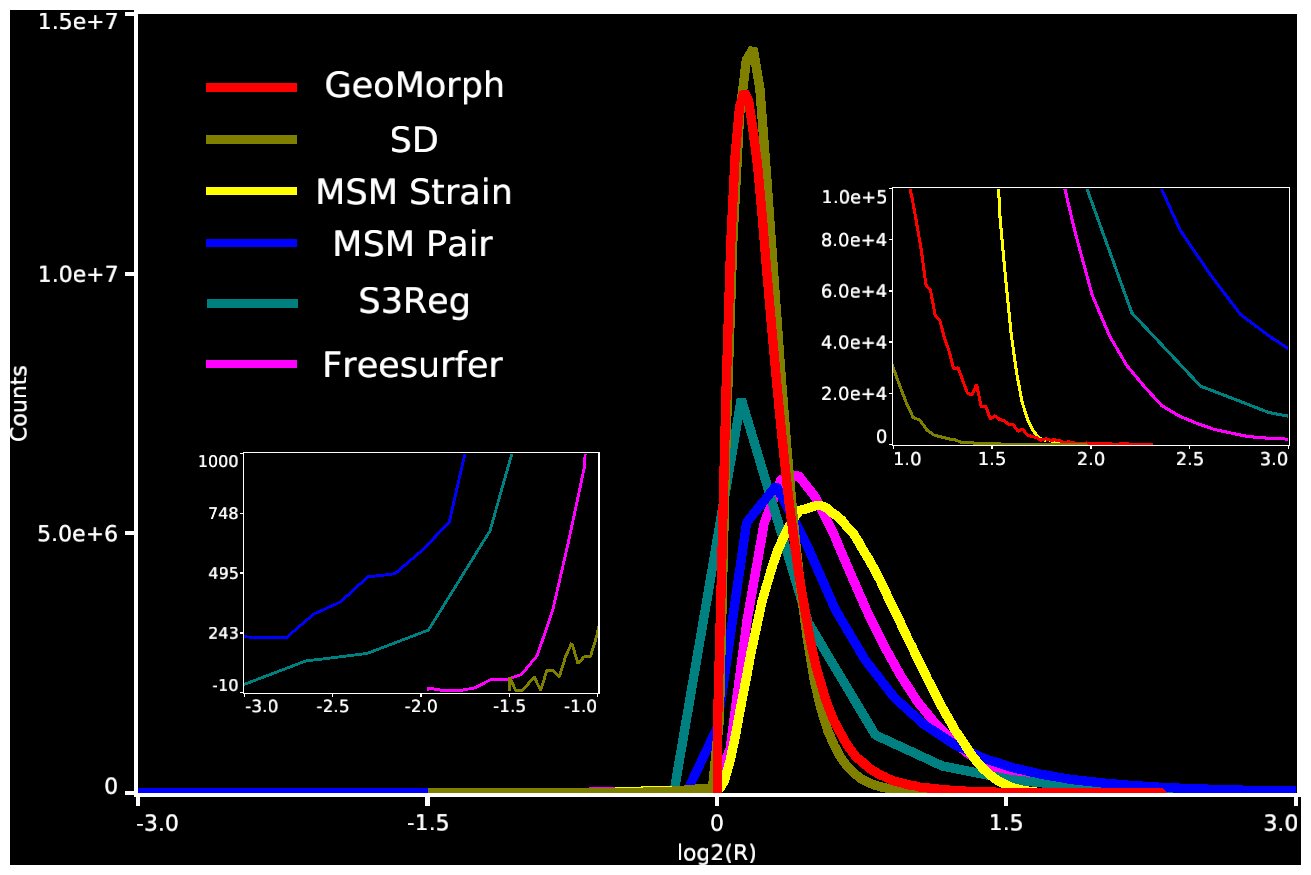}}
  }
   {\caption{Histogram plots comparing areal and shape distortions across all test subjects.} \label{fig:distortions}} 
\end{figure}

In Fig.~\ref{fig:distortions}, we present histograms showcasing the distribution of both areal and shape distortions, across all test subjects, at CC of 0.88. By analyzing Fig.~\ref{fig:distortions}a, we observe that SD and GeoMorph predominantly exhibit areal distortions centered around zero. 
On the contrary, MSM Pair, S3Reg, and (to a lesser degree) Freesurfer demonstrate pronounced extreme distortions across subjects, as evidenced by the presence of long tails in the histogram. This trend is also observable in Fig.~\ref{fig:distortions}b, where we observe that the distribution of shape distortions, for SD and GeoMorph, predominantly falls below one. Conversely, other methods exhibit significant instances of extreme distortions.

\begin{figure*}[h!]
\centering
  {%
    \subfigure[]{\label{fig:aveg_perf}%
      \includegraphics[width=7. in]{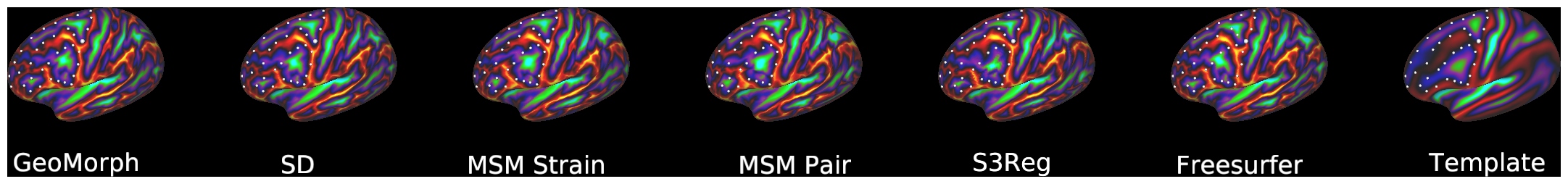}}%
    \qquad
    \subfigure[]{\label{fig:single_dists}%
      \includegraphics[width=7. in]{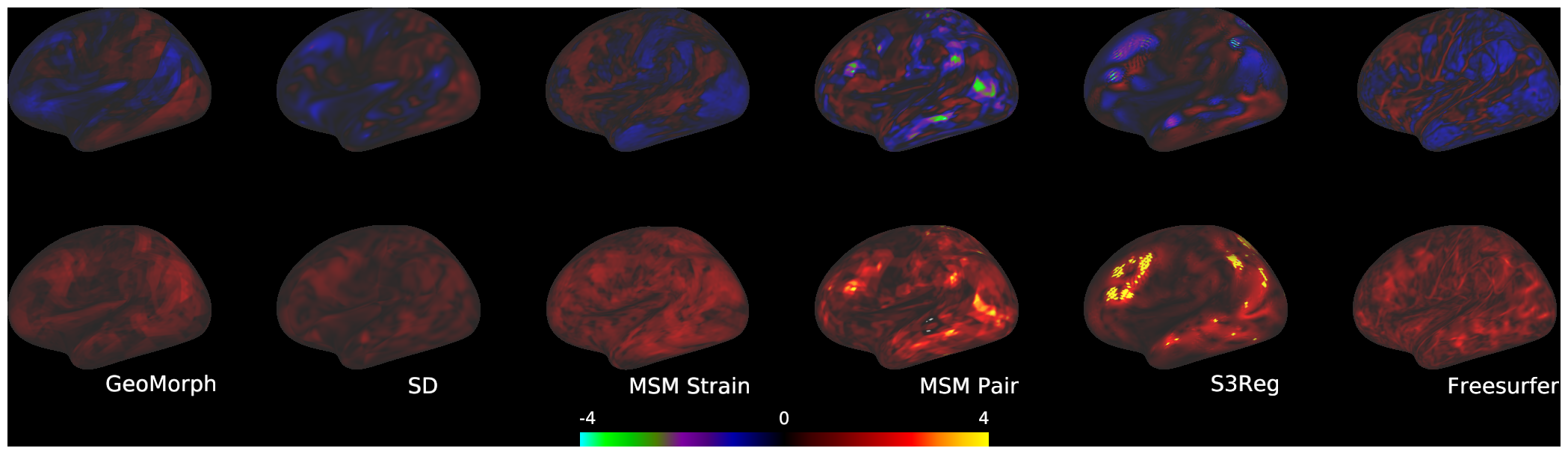}}
  }
   {\caption{a) Registration Performance. b) Areal (top) and shape (bottom) distortions.} \label{fig:single_subject_per}} 
\end{figure*}

In Fig.~\ref{fig:single_subject_per}, we visually evaluate alignment quality, for all benchmarked methods, on one subject from the HCP dataset that exhibits atypical cortical folding patterns. Additionally, we examine the areal and shape distortions resulting from these methods. The figure illustrates that GeoMorph, SD, and MSM Strain achieve favorable alignment, with minimal distortions. In contrast, the alignments produced by MSM Pair and S3Reg are characterized by regions with very high distortions.

\begin{table*}[!ht]
\begin{center}
\setlength{\tabcolsep}{1.6pt}
\caption{Mutimodal registration results: CC, distortions measures, and average runtime for different methods. HCP (top) and UKB (bottom).}\label{table: multimodal results}
\centering
    \begin{tabular}{cc|c||c|c||c|c|c|c||c|c|c|c||c|c|}
        \hline
        \hline
        \multicolumn{1}{|c|}{\multirow{2}{*}{\textbf{Methods}}}          & \multicolumn{2}{|c||}{\textbf{CC Similarity}}  & \multicolumn{2}{|c||}{\textbf{Cluster mass}}  & \multicolumn{4}{|c||}{\textbf{Areal Distortion}}  & \multicolumn{4}{c||}{\textbf{Shape Distortion}} & \multicolumn{2}{c|}{\textbf{Avg. Time}}  \\ \cline{2-15}
        \multicolumn{1}{|c|}{}            & \textbf{Myelin}   &\textbf{rfMRI}    & \textbf{Myelin}   &\textbf{rfMRI}  &\textbf{Mean}   & \textbf{Max}       & \textbf{95\%} & \textbf{98\%}                           &\textbf{Mean }      &\textbf{Max}       & \textbf{95\%}  & \textbf{98\%}  & \textbf{CPU}         &\textbf{GPU}  \\ \hline
        \hline
        \multicolumn{1}{|c|}{\begin{tabular}[c]{@{}c@{}}  MSMSulc \end{tabular}}   
        
              &0.938    &0.523  &5269  &130800 &0.1 &0.35 & 0.16   & 0.18 & 0.21  &0.64  &0.38   & 0.42   & 40 min   & -   \\                            
                    
        \multicolumn{1}{|c|}{\begin{tabular}[c]{@{}c@{}} GeoMorphSulc \end{tabular}}
        
         & 0.954     &  0.523 &5382 &133575 & 0.19  &2.43  &0.53  & 0.69  &0.26  &2.70    &0.63   &0.82    & 8.3 sec&   2.6 sec     \\ 
     
     \multicolumn{1}{|c|}{\begin{tabular}[c]{@{}c@{}} MSMAll \end{tabular}}
        
         &0.945  & 0.566    &5546  &182636  & 0.27   &1.2  & 0.62       & 0.68    & 0.62   &1.92  & 1.13     & 1.24     &  1.5 hour   & -        \\ 
         
         \multicolumn{1}{|c|}{\begin{tabular}[c]{@{}c@{}}  GeoMorphAll   \end{tabular}}   
        
        &\textbf{0.975}  &\textbf{0.569}  &\textbf{5797}  &\textbf{182734}   & 0.24   &  7.57  &0.64   &0.84   &0.35   &8.5  &0.82  &1.0   & \textbf{7.7 sec} &  \textbf{0.55 sec }    \\ \hline \hline
        
        \multicolumn{1}{|c|}{\begin{tabular}[c]{@{}c@{}}  MSMSulc \end{tabular}}   
        
              &  0.75   & 0.35 &96476  &71105 &0.34  & 0.1 & 0.35     &   0.17 & 0.23  & 0.65  & 0.41   &0.45   & 40 min   & -   \\                            
                    
        \multicolumn{1}{|c|}{\begin{tabular}[c]{@{}c@{}} GeoMorphSulc \end{tabular}}
        
         & 0.94     &  0.36  &96591   &80973  & 0.19 &2.43  &0.52  & 0.68  &0.30  &3.20    &0.68   &0.87    & 8.3 sec&   2.6 sec     \\

     \multicolumn{1}{|c|}{\begin{tabular}[c]{@{}c@{}} MSMAll \end{tabular}}
        
         & 0.944  &\textbf{0.40}    &96779  &\textbf{105866} & 0.27  &2.4  & 0.63      & 0.70    &0.66   &3.10  &1.2  &1.40   &  1.5 hour   & -        \\ 
         
         \multicolumn{1}{|c|}{\begin{tabular}[c]{@{}c@{}}  GeoMorphAll   \end{tabular}}   
        
        & \textbf{0.96} &\textbf{0.40}  &\textbf{97319}  &103358  & 0.28 &  7.72  &0.86   &1.21  &0.41  &8.78  &1.0  &1.53   & \textbf{7.7 sec} &  \textbf{0.55 sec}   \\ \hline \hline
    \end{tabular} 
    \end{center}  
\end{table*}

%

\subsection{Multimodal Registration}
\label{sec: multimodal registration}
\begin{figure*}[ht!]
\centering
\includegraphics[width=6.5 in]{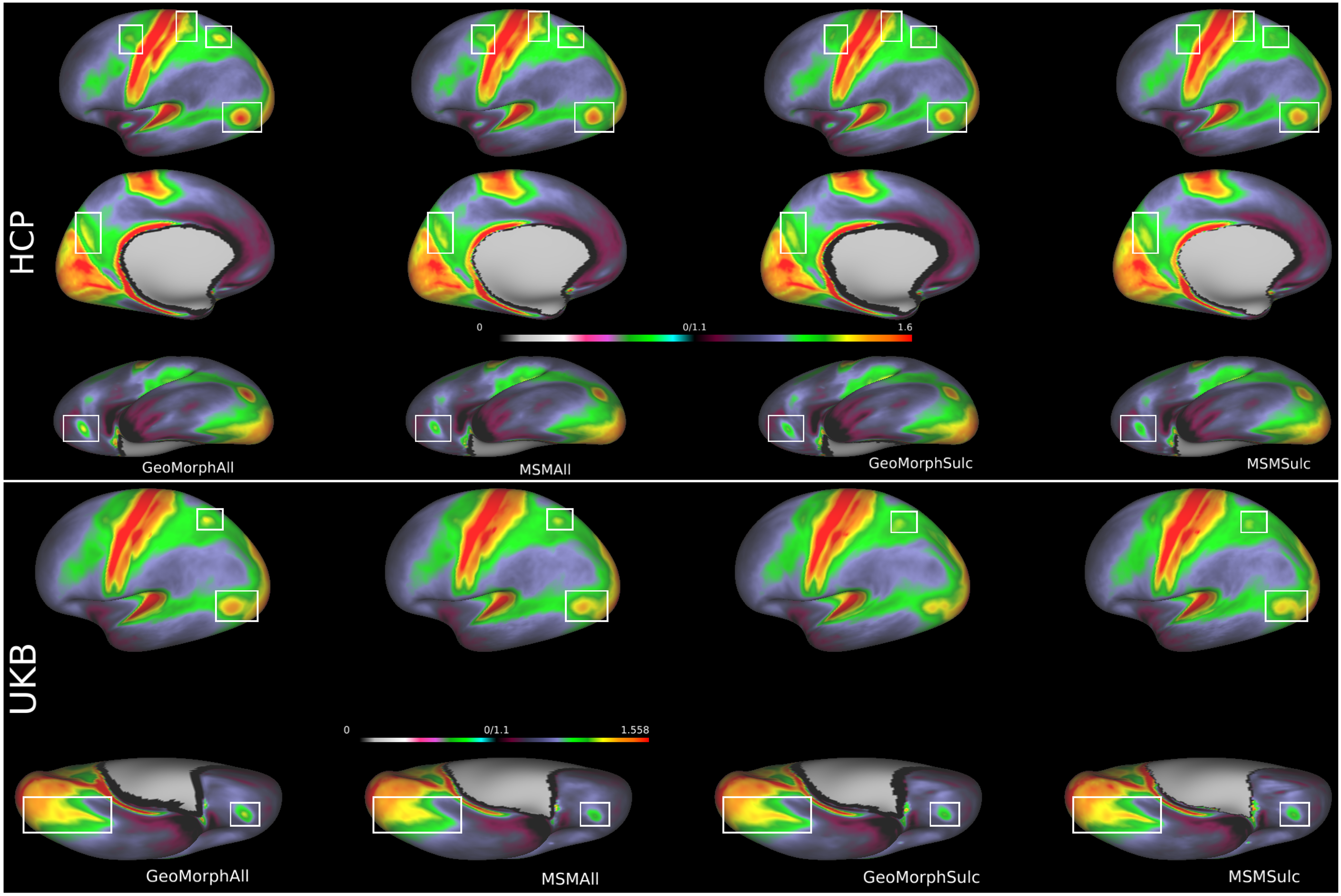}
\caption{Intersubject averages of myelin features in HCP (top) and UKB (bottom). Regions in the brain that exhibit significant disparities in contrasts and enhanced performance between these methods are denoted by the areas highlighted within white boxes.} \label{fig: myelin}
\end{figure*}

Fig.~\ref{fig: myelin} illustrates the sharpness of cross-subject T1w/T2w myelin map averages for all methods, benchmarked for both HCP and UKB datasets. Areas of distinct improvement, following multimodal alignment, are highlighted by white boxes. The results show that GeoMorphAll yields a sharper and clearer average map, relative to MSMAll. Improvements are particularly distinct for the UKB dataset; this might be because MSM was originally optimized for the HCP. 

\begin{figure*}[ht!]
\centering
\includegraphics[width=6.5 in]{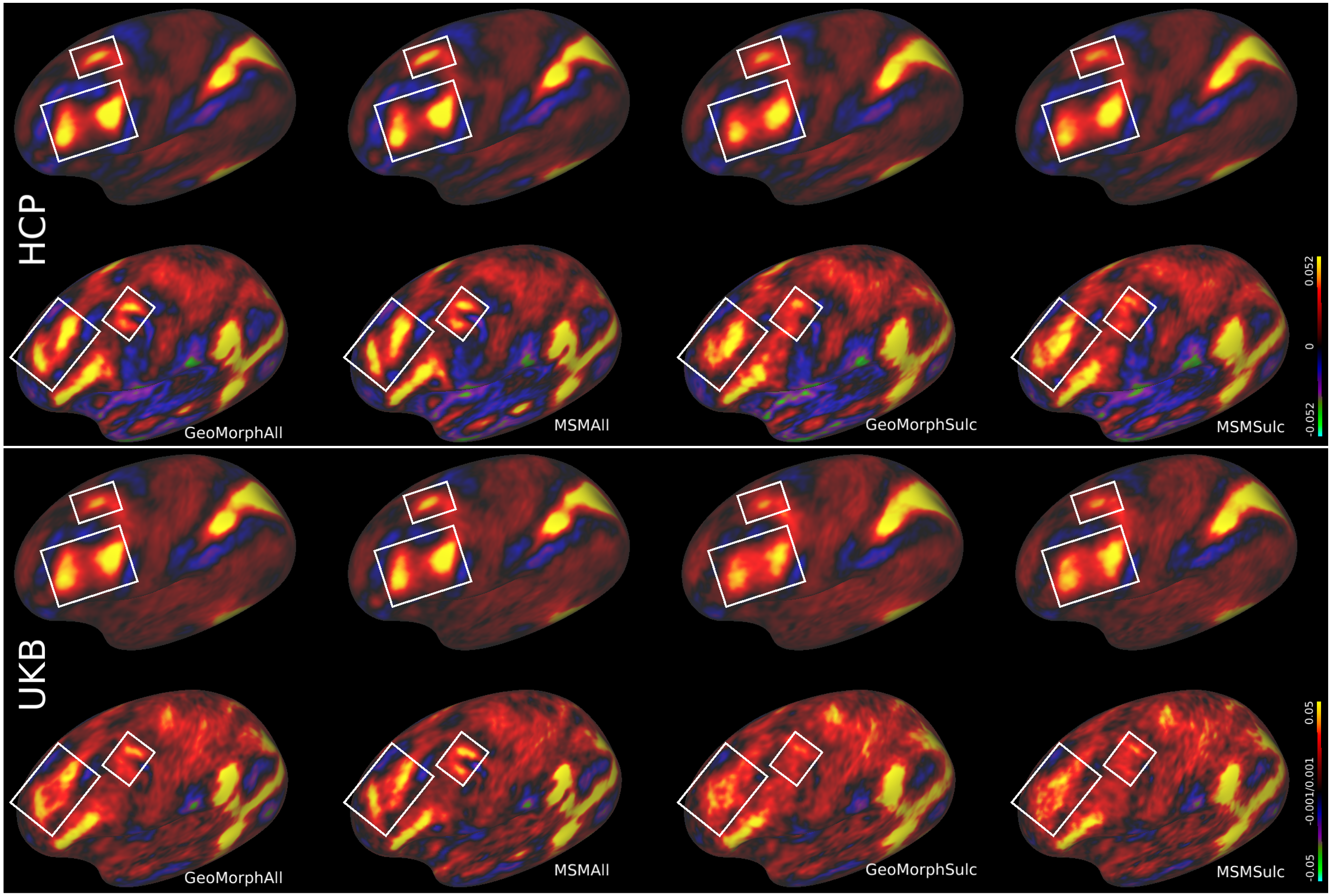}
\caption{Intersubject averages of rfMRI features in HCP (top) and UKB (bottom). Regions in the brain that exhibit significant disparities in contrasts and enhanced performance between these methods are denoted by the areas highlighted within white boxes.} \label{fig: rsns}
\end{figure*}

Similar patterns of results are shown for the cross-subject averages of resting state spatial maps (Fig.~\ref{fig: rsns}). In this case performance relative to MSM is more variable across the cortex; however, the improvements of GeoMorphAll and MSMAll, relative to GeoMorphSulc and MSMSulc, remain clear.

Table.~\ref{table: multimodal results} provides a quantitative analysis of the registration performance, on myelin and rfMRI, in terms of CC similarity and distortion statistics. The results reveal that GeoMorphAll demonstrates comparable CC performance to MSMAll, across both datasets. Notably, while MSMSulc exhibits superior distortion measures, this is to be expected as it reflects a highly constrained folding based alignment. It is clear this comes at the expense of much poorer CC performance (Figs.~\ref{fig: myelin} and \ref{fig: rsns}).
 
\begin{figure*}[h]
\centering
\includegraphics[width=6. in]{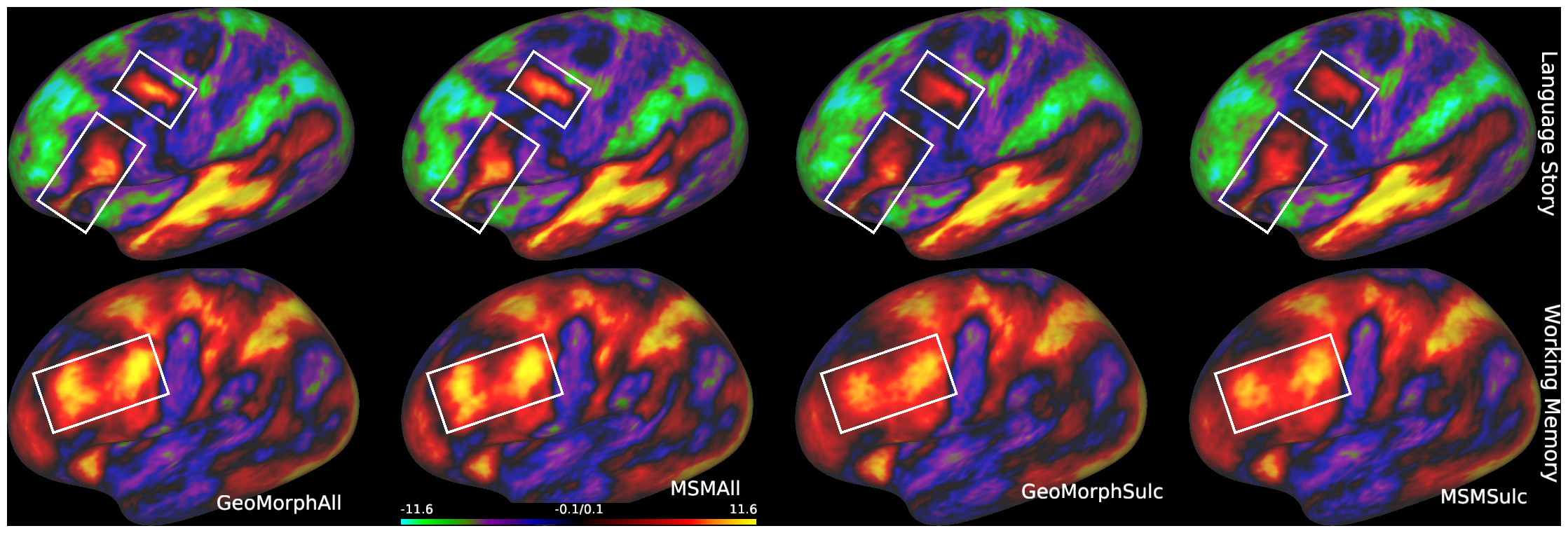}
\caption{Comparison of group Z-statistic spatial maps of all methods for a) Language Story task. b) Working Memory task. White boxes highlight improvements in the sharpness of the contrast, with GeoMorphAll and MSMAll maintaining a close comparable performance that outperforms unimodal methods.} \label{fig: task_results}
\vspace{-10pt}
\end{figure*}

Validation on independently acquired tfMRI is reported in Table.~\ref{table: cluster mass}. Results reproduce previous findings from MSM studies \cite{coalson2018impact,glasser2016human, glasser2016multi,robinson2014msm,robinson2018multimodal,smith2013functional} that suggest that multimodal alignment significantly improves the overlap of cortical functional areas across subjects, as evidenced by improvements to the extent and peak z-values of group task statistics (as summarised by the cluster mass measure). Overall performance of GeoMorphAll is comparable to that of MSMAll. Fig.~\ref{fig: task_results} presents visual results for the working memory and language story task. The figure shows the improvement in the sharpness of the contrast in multiple areas within the brain for both GeoMorphAll and MSMAll, both outperforming unimodal alignment methods using sulcal depth.

\begin{table*}[h]
\begin{center}
\caption{Cluster mass estimates of the aligned HCP tfMRI data using proposed methods.}\label{table: cluster mass}
\centering
    \begin{tabular}{cc|c|c|c|}
        \hline
        \hline
        \multicolumn{1}{|c|}{\multirow{2}{*}{\textbf{Task Name}}}   
                    
     & \multicolumn{1}{|c|}{\multirow{2}{*}{\textbf{MSMSulc}}}  
     
      & \multicolumn{1}{|c|}{\multirow{2}{*}{\textbf{GeoMorphSulc}}} 
          
     & \multicolumn{1}{|c|}{\multirow{2}{*}{\textbf{MSMAll}}}  
        
     & \multicolumn{1}{|c|}{\multirow{2}{*}{\textbf{GeoMorphAll}}} 
        
        \\ 
        
        \multicolumn{1}{|c|}{}  &   &  &   &\\ \hline
        \hline
        
        \multicolumn{1}{|c|}{\begin{tabular}[c]{@{}c@{}}  Emotion  \end{tabular}}   
        
              &251677     & 248641  & \textbf{303403}  &290406\\                            
                    
        \multicolumn{1}{|c|}{\begin{tabular}[c]{@{}c@{}} Gambling \end{tabular}}
        
         &352831      &350187  &\textbf{418681} & 405669\\ 
     
     \multicolumn{1}{|c|}{\begin{tabular}[c]{@{}c@{}} Relational Processing \end{tabular}}
        
         &550403     &552769  & \textbf{611565}   &602924\\ 
         
         \multicolumn{1}{|c|}{\begin{tabular}[c]{@{}c@{}}  Language Story \end{tabular}}   
        
        &558434     &  570587  & 634654 & \textbf{654618} \\ 
        
        \multicolumn{1}{|c|}{\begin{tabular}[c]{@{}c@{}} Social Cognition \end{tabular}}
        
         & 625291     & 618237  &  687310  &\textbf{690586} \\ 
         
         \multicolumn{1}{|c|}{\begin{tabular}[c]{@{}c@{}} Motor \end{tabular}}
        
         &1485991     &1487278  &\textbf{1617727}  &1612465 \\ 
         
         \multicolumn{1}{|c|}{\begin{tabular}[c]{@{}c@{}} Working Memory \end{tabular}}
        
         & 1961989    &1936564  &\textbf{2240070}  &2190214 \\  \hline \hline
         
    \end{tabular} 
    \end{center}  
\end{table*}

\subsection{Ablation and hyperparameter tuning of multimodal alignment}
\label{sec:ablation}
\begin{figure}[h]
\begin{center}
\includegraphics[width=2.5 in]{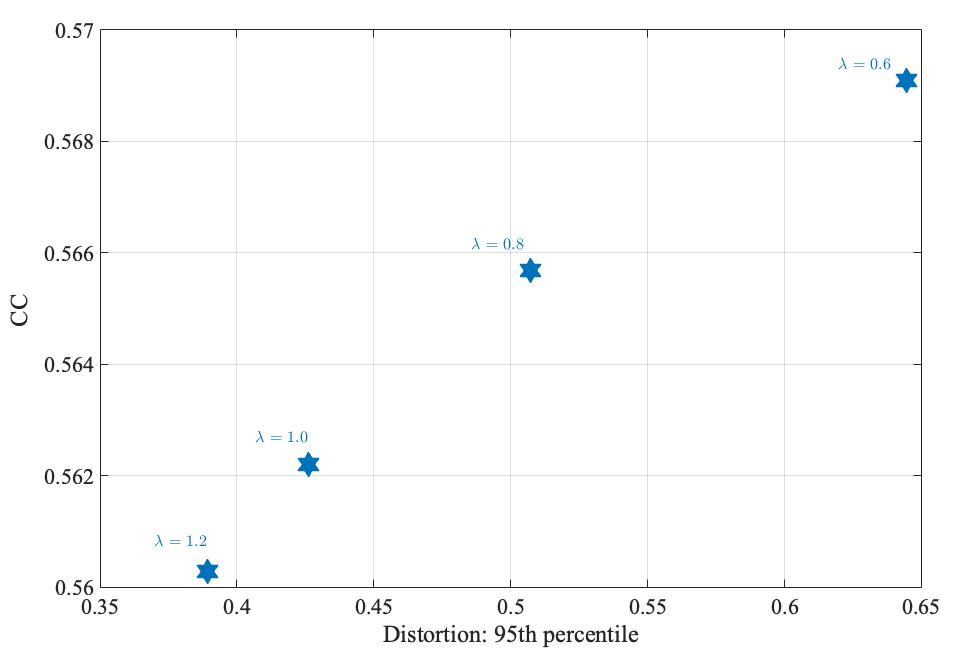}
\caption{Similarity performances vs. the 95th percentile of the areal distortion at multiple regularization levels} \label{fig: per vs runs}
\end{center}
\vspace{-15pt}
\end{figure}

Fig.~\ref{fig: per vs runs}, reports performance of the network as the regularisation parameter $\lambda$ is changed, indicating an inverse relationship between image similarity and distortion. 
The impact of control point grid resolution is reported in Table.~\ref{table: ablation control resolution}. These results demonstrate that as the resolution of the control grid increases, the model's performance improves.

\begin{table}[h]
\begin{center}
\caption{Model performance at multiple control points resolutions.}\label{table: ablation control resolution}
\centering
    \begin{tabular}{cc|c|c|}
        \hline
        \hline
        \multicolumn{1}{|c|}{\multirow{2}{*}{\textbf{Measure}}}   
                    
     & \multicolumn{1}{|c|}{\multirow{2}{*}{\textbf{ico-2}}}  
     
      & \multicolumn{1}{|c|}{\multirow{2}{*}{\textbf{ico-3}}} 
      
      & \multicolumn{1}{|c|}{\multirow{2}{*}{\textbf{ico-4}}} 
        \\ 
        
        \multicolumn{1}{|c|}{}  &   &   &\\ \hline
        \hline
        \multicolumn{1}{|c|}{\begin{tabular}[c]{@{}c@{}}  CC Similarity: Myelin\end{tabular}}   
        
              & 0.969    & 0.972 & \textbf{0.975}\\ 
              
        \multicolumn{1}{|c|}{\begin{tabular}[c]{@{}c@{}}  CC Similarity: rfMRI\end{tabular}}   
        
              & 0.563    & 0.568  & \textbf{0.569}\\

        \multicolumn{1}{|c|}{\begin{tabular}[c]{@{}c@{}}  Distortion: 95\%  \end{tabular}}   
        
              &\textbf{0.55 }    & 0.60   & 0.64 \\                            
                    
        \multicolumn{1}{|c|}{\begin{tabular}[c]{@{}c@{}} CM: Myelin \end{tabular}}
        
         &5657    & 5778  & \textbf{5797} \\ 
         
         \multicolumn{1}{|c|}{\begin{tabular}[c]{@{}c@{}} CM: rfMRI \end{tabular}}
        
         &175967   & 182510  & \textbf{182734} \\

     \multicolumn{1}{|c|}{\begin{tabular}[c]{@{}c@{}} CM: Language Story \end{tabular}}
        
         &641751    &649699  &\textbf{654618}\\

          \multicolumn{1}{|c|}{\begin{tabular}[c]{@{}c@{}} CM: Motor \end{tabular}}
        
         &1591248     &1605115  & \textbf{1612465}\\  \hline \hline    
    \end{tabular} 
    \end{center}  
    \vspace{-15pt}
\end{table}

Ablation analyses, investigating the benefits of different components of the network are summarised in Table~\ref{table: CRF-RNN results}.  
These demonstrate that integration of a CFR-RNN network, reduces distortion (across all measures) whilst not impacting the goodness of alignment (as quantified through the CC similarity measure).

\begin{table*}[!h]
\begin{center}
\caption{Comparisons of GeoMorphAll registration results with/without CRF-RNN regularization network: CC similarity and  distortions measures. HCP (top) and UKB (bottom).}\label{table: CRF-RNN results}
\centering
    \begin{tabular}{cc|c||c|c|c|c||c|c|c|c||}
        \hline
        \hline
        \multicolumn{1}{|c|}{\multirow{2}{*}{\textbf{CRF-RNN}}}    &  \multicolumn{2}{|c||}{\textbf{CC Similarity}}  & \multicolumn{4}{|c||}{\textbf{Areal Distortion}}  & \multicolumn{4}{c||}{\textbf{Shape Distortion}}  \\ \cline{2-11}
         \multicolumn{1}{|c|}{}          & \textbf{Myelin}   &\textbf{rfMRI}    &\textbf{Mean}   & \textbf{Max}       & \textbf{95\%} & \textbf{98\%}                           &\textbf{Mean }      &\textbf{Max}       & \textbf{95\%}  & \textbf{98\%}  \\ \hline
        \hline     

         \multicolumn{1}{|c|}{\begin{tabular}[c]{@{}c@{}}  \cmark   \end{tabular}}   
        
        &0.975  &0.569    &\textbf{0.24 }  &  \textbf{7.57} &\textbf{0.64}  &0\textbf{.84}   &\textbf{0.35}   &\textbf{8.5}  &\textbf{0.82} &\textbf{1.0}    \\

        \multicolumn{1}{|c|}{\begin{tabular}[c]{@{}c@{}} \xmark \end{tabular}}
        
          & 0.976    &0.569  & 0.26  &8.64  &0.71 & 0.91  &0.39  &9.4    &0.86   &1.32      \\ 
         
        \hline \hline
                
         \multicolumn{1}{|c|}{\begin{tabular}[c]{@{}c@{}}  \cmark    \end{tabular}}   
        
        & 0.96 &0.40  & \textbf{0.28} &  \textbf{7.72}  &\textbf{0.86 }  &\textbf{1.21}  &\textbf{0.41}  &\textbf{8.78}  &\textbf{1.0}  &\textbf{1.53}     \\ 
        
        \multicolumn{1}{|c|}{\begin{tabular}[c]{@{}c@{}} \xmark \end{tabular}}
        
       & 0.958     & 0.404   & 0.31  &11.44  &0.94  & 1.69  &0.49  &11.5    &1.31   &1.84    \\ 
         
        \hline \hline
    \end{tabular} 
    \end{center} 
    \vspace{-10pt}
\end{table*}

\section{Conclusions and Discussions}
\label{sec: conclusion}
In this paper, we presented GeoMorph, an innovative geometric deep-learning framework specifically designed for multimodal cortical surface registration. GeoMorph aims to learn a smooth displacement field that effectively aligns the features on the moving surface with those on the target surface. By leveraging independent feature extraction and deep-discrete registration, GeoMorph captures crucial characteristics of cortical surfaces and optimizes feature overlap to achieve improved alignment. To ensure the generation of visually coherent and anatomically plausible deformations, we incorporate a regularization network based on a deep conditional random field. Experimental results using sulcal depth features demonstrate that GeoMorph outperforms existing deep-learning methods by achieving enhanced alignment and generating smoother deformations. Moreover, GeoMorph exhibits competitive performance when compared to classical frameworks while demonstrating efficient run times. This practicality makes GeoMorph a suitable choice for real-world applications. Furthermore, in the context of multimodal registration, which incorporates myelin and rfMRI features, GeoMorph demonstrates superior individual-to-template alignment when evaluated on these two features. When compared to MSMAll, the leading classical multimodal registration framework, GeoMorphAll exhibits improved performance on both the HCP and the UKB datasets while significantly reducing the execution run time.

The results in Section~\ref{sec: results} reveal that GeoMorph demonstrates comparable performance to the classical MSM approach in most cases. However, GeoMorph gains a significant advantage from its faster execution time (as demonstrated in Tables \ref{table: dist} and \ref{table: multimodal results}), making it computationally efficient in learning population-specific templates. These templates capture crucial attribute-related trends, which sets GeoMorph apart from conventional methods that only generate population average templates capturing dominant folding patterns \cite{dalca2019learning}. This computational efficiency proves particularly beneficial for clinical applications, especially when a pre-existing template does not exist. In such cases, using methods like MSM to obtain a template would be prohibitively expensive. GeoMorph's ability to efficiently generate population-specific templates makes it a highly valuable choice for these scenarios.

When comparing the performance of GeoMorph against the deep-learning based method S3Reg (as shown in Table~\ref{table: dist}), it becomes evident that S3Reg exhibits significant high peak distortions. This issue is likely attributed to the hexagonal filter utilized in S3Reg, which lacks rotational equivariance due to the absence of a global spherical coordinate system. The solution proposed by S3Reg, which involves a combination of three networks, does not seem to fully overcome this problem. Furthermore, S3Reg attempts to enforce diffeomorphisms by using the scaling and squaring approach from the diffeomorphic Voxelmorph algorithm \cite{dalca2019learning}. Contrastingly, the GeoMorph structure does not explicitly prioritize the enforcement of diffeomorphisms. Nevertheless, all yielded results are discovered to be diffeomorphic—a foreseeable outcome due to the strong regularization imposed by the GRF-RNN network, which effectively ensures this stipulation. A future expansion of GeoMorph aims to explore relaxing this requirement within the framework. The motivation for delving into this aspect arises from recognizing that achieving perfect alignment on functional topographies proves unattainable with diffeomorphically-constrained deformations. This observation is evident in 10\% of subjects for area 55b in \cite{glasser2016multi}, as underscored in \cite{schneider2019columnar}. Conventional diffeomorphic registration approaches, as showcased in \cite{thual2022aligning}, tend to exhibit reduced performance in such scenarios.

Section~\ref{sec: multimodal registration} presents compelling evidence of the advantages offered by GeoMorphAll in achieving enhanced registration outcomes. The inclusion of myelin and rfMRI features in the registration process has proven to be beneficial for individual-to-template alignment. A comparison between GeoMorphAll and MSMAll results, as demonstrated in Figs~\ref{fig: myelin} and ~\ref{fig: rsns} using both HCP and UKB datasets, reveals that GeoMorphAll produces a sharper and clearer average map in the UKB dataset when compared to MSMAll. This difference may arise from the fact that MSMAll was originally optimized for the HCP dataset and that the feature extraction network in GeoMorph is more advantageous in the context of the UKB dataset, where the data is considerably noisier compared to the HCP dataset.

The computational complexity of GeoMorph is heavily dependent on the resolution of the control points. As a result, we have been limited to a control grid on an icosphere of level 4 due to memory constraints. Hence, future extensions of this work will explore more efficient methods to overcome this limitation. Additionally, while the CRF has proven to be beneficial in regularizing the deformation field, there is a keen interest in investigating other regularization techniques that fully leverage the characteristics of the problem. One promising approach is the development of a mechanically learnable regularization penalty that considers the physical properties of brain tissues. This approach aims to obtain improved and robust deformations while also allowing for topology-breaking transformations. Furthermore, it's worth noting that GeoMorph's implementation is entirely based on MoNet convolutions that somewhat limit the expressivity of the features derived from the network \cite{fawaz2021benchmarking}. Improved performance might be achievable through spectral learning frameworks such as S2CNN \cite{cohen2018spherical}, which learn fully expressive, rotation-equivariant convolutions for the sphere, or through incorporation of surface transformer networks \cite{dahan2022surface}. However, doing so would require new solutions that improve the computational efficiency of these networks. 
\section*{Acknowledgments}
Data were provided [in part] by the Human Connectome Project, WU-Minn Consortium (Principal Investigators: David Van Essen and Kamil Ugurbil; 1U54MH091657) funded by the 16 NIH Institutes and Centers that support the NIH Blueprint for Neuroscience Research; and by the McDonnell Center for Systems Neuroscience at Washington University.

UK Biobank data were accessed under application number 53775, Principal Investigator Dr. Emma C. Robinson.


\end{document}

%% file: macros.tex
\setlength\unitlength{1mm}

\newcommand{\insertfig}[3]{
\begin{figure}[htbp]\begin{center}\begin{picture}(120,90)
\put(0,-5){\includegraphics[width=12cm,height=9cm,clip=]{#1.eps}}\end{picture}\end{center}
\caption{#2}\label{#3}\end{figure}}

\newcommand{\insertxfig}[4]{
\begin{figure}[htbp]
\begin{center}
\leavevmode \centerline{\resizebox{#4\textwidth}{!}{\input
#1.pstex_t}}
\caption{#2} \label{#3}
\end{center}
\end{figure}}

\long\def\comment#1{}



\newfont{\bbb}{msbm10 scaled 700}
\newcommand{\CCC}{\mbox{\bbb C}}

\newfont{\bb}{msbm10 scaled 1100}
\newcommand{\CC}{\mbox{\bb C}}
\newcommand{\PP}{\mbox{\bb P}}
\newcommand{\RR}{\mbox{\bb R}}
\newcommand{\QQ}{\mbox{\bb Q}}
\newcommand{\ZZ}{\mbox{\bb Z}}
\newcommand{\FF}{\mbox{\bb F}}
\newcommand{\GG}{\mbox{\bb G}}
\newcommand{\EE}{\mbox{\bb E}}
\newcommand{\NN}{\mbox{\bb N}}
\newcommand{\KK}{\mbox{\bb K}}


\newcommand{\av}{{\bf a}}
\newcommand{\bv}{{\bf b}}
\newcommand{\cv}{{\bf c}}
\newcommand{\dv}{{\bf d}}
\newcommand{\ev}{{\bf e}}
\newcommand{\fv}{{\bf f}}
\newcommand{\gv}{{\bf g}}
\newcommand{\hv}{{\bf h}}
\newcommand{\iv}{{\bf i}}
\newcommand{\jv}{{\bf j}}
\newcommand{\kv}{{\bf k}}
\newcommand{\lv}{{\bf l}}
\newcommand{\mv}{{\bf m}}
\newcommand{\nv}{{\bf n}}
\newcommand{\ov}{{\bf o}}
\newcommand{\pv}{{\bf p}}
\newcommand{\qv}{{\bf q}}
\newcommand{\rv}{{\bf r}}
\newcommand{\sv}{{\bf s}}
\newcommand{\tv}{{\bf t}}
\newcommand{\uv}{{\bf u}}
\newcommand{\wv}{{\bf w}}
\newcommand{\vv}{{\bf v}}
\newcommand{\xv}{{\bf x}}
\newcommand{\yv}{{\bf y}}
\newcommand{\zv}{{\bf z}}
\newcommand{\zerov}{{\bf 0}}
\newcommand{\onev}{{\bf 1}}

\def\u{{\bf u}}


\newcommand{\Am}{{\bf A}}
\newcommand{\Bm}{{\bf B}}
\newcommand{\Cm}{{\bf C}}
\newcommand{\Dm}{{\bf D}}
\newcommand{\Em}{{\bf E}}
\newcommand{\Fm}{{\bf F}}
\newcommand{\Gm}{{\bf G}}
\newcommand{\Hm}{{\bf H}}
\newcommand{\Id}{{\bf I}}
\newcommand{\Jm}{{\bf J}}
\newcommand{\Km}{{\bf K}}
\newcommand{\Lm}{{\bf L}}
\newcommand{\Mm}{{\bf M}}
\newcommand{\Nm}{{\bf N}}
\newcommand{\Om}{{\bf O}}
\newcommand{\Pm}{{\bf P}}
\newcommand{\Qm}{{\bf Q}}
\newcommand{\Rm}{{\bf R}}
\newcommand{\Sm}{{\bf S}}
\newcommand{\Tm}{{\bf T}}
\newcommand{\Um}{{\bf U}}
\newcommand{\Wm}{{\bf W}}
\newcommand{\Vm}{{\bf V}}
\newcommand{\Xm}{{\bf X}}
\newcommand{\Ym}{{\bf Y}}
\newcommand{\Zm}{{\bf Z}}
\newcommand{\Onem}{{\bf 1}}
\newcommand{\Zerom}{{\bf 0}}


\newcommand{\Ac}{{\cal A}}
\newcommand{\Bc}{{\cal B}}
\newcommand{\Cc}{{\cal C}}
\newcommand{\Dc}{{\cal D}}
\newcommand{\Ec}{{\cal E}}
\newcommand{\Fc}{{\cal F}}
\newcommand{\Gc}{{\cal G}}
\newcommand{\Hc}{{\cal H}}
\newcommand{\Ic}{{\cal I}}
\newcommand{\Jc}{{\cal J}}
\newcommand{\Kc}{{\cal K}}
\newcommand{\Lc}{{\cal L}}
\newcommand{\Mc}{{\cal M}}
\newcommand{\Nc}{{\cal N}}
\newcommand{\Oc}{{\cal O}}
\newcommand{\Pc}{{\cal P}}
\newcommand{\Qc}{{\cal Q}}
\newcommand{\Rc}{{\cal R}}
\newcommand{\Sc}{{\cal S}}
\newcommand{\Tc}{{\cal T}}
\newcommand{\Uc}{{\cal U}}
\newcommand{\Wc}{{\cal W}}
\newcommand{\Vc}{{\cal V}}
\newcommand{\Xc}{{\cal X}}
\newcommand{\Yc}{{\cal Y}}
\newcommand{\Zc}{{\cal Z}}


\newcommand{\alphav}{\hbox{\boldmath$\alpha$}}
\newcommand{\betav}{\hbox{\boldmath$\beta$}}
\newcommand{\gammav}{\hbox{\boldmath$\gamma$}}
\newcommand{\deltav}{\hbox{\boldmath$\delta$}}
\newcommand{\etav}{\hbox{\boldmath$\eta$}}
\newcommand{\lambdav}{\hbox{\boldmath$\lambda$}}
\newcommand{\epsilonv}{\hbox{\boldmath$\epsilon$}}
\newcommand{\nuv}{\hbox{\boldmath$\nu$}}
\newcommand{\muv}{\hbox{\boldmath$\mu$}}
\newcommand{\zetav}{\hbox{\boldmath$\zeta$}}
\newcommand{\phiv}{\hbox{\boldmath$\phi$}}
\newcommand{\psiv}{\hbox{\boldmath$\psi$}}
\newcommand{\thetav}{\hbox{\boldmath$\theta$}}
\newcommand{\tauv}{\hbox{\boldmath$\tau$}}
\newcommand{\omegav}{\hbox{\boldmath$\omega$}}
\newcommand{\xiv}{\hbox{\boldmath$\xi$}}
\newcommand{\sigmav}{\hbox{\boldmath$\sigma$}}
\newcommand{\piv}{\hbox{\boldmath$\pi$}}
\newcommand{\rhov}{\hbox{\boldmath$\rho$}}
\newcommand{\vtv}{\hbox{\boldmath$\vartheta$}}

\newcommand{\Gammam}{\hbox{\boldmath$\Gamma$}}
\newcommand{\Lambdam}{\hbox{\boldmath$\Lambda$}}
\newcommand{\Deltam}{\hbox{\boldmath$\Delta$}}
\newcommand{\Sigmam}{\hbox{\boldmath$\Sigma$}}
\newcommand{\Phim}{\hbox{\boldmath$\Phi$}}
\newcommand{\Pim}{\hbox{\boldmath$\Pi$}}
\newcommand{\Psim}{\hbox{\boldmath$\Psi$}}
\newcommand{\psim}{\hbox{\boldmath$\psi$}}
\newcommand{\chim}{\hbox{\boldmath$\chi$}}
\newcommand{\omegam}{\hbox{\boldmath$\omega$}}
\newcommand{\vphim}{\hbox{\boldmath$\varphi$}}
\newcommand{\Thetam}{\hbox{\boldmath$\Theta$}}
\newcommand{\Omegam}{\hbox{\boldmath$\Omega$}}
\newcommand{\Xim}{\hbox{\boldmath$\Xi$}}


\newcommand{\sinc}{{\hbox{sinc}}}
\newcommand{\diag}{{\hbox{diag}}}
\renewcommand{\det}{{\hbox{det}}}
\newcommand{\trace}{{\hbox{tr}}}
\newcommand{\sign}{{\hbox{sign}}}
\renewcommand{\arg}{{\hbox{arg}}}
\newcommand{\var}{{\hbox{var}}}
\newcommand{\cov}{{\hbox{cov}}}
\newcommand{\SINR}{{\sf sinr}}
\newcommand{\SNR}{{\sf snr}}
\newcommand{\Ei}{{\rm E}_{\rm i}}
\newcommand{\eqdef}{\stackrel{\Delta}{=}}
\newcommand{\defines}{{\,\,\stackrel{\scriptscriptstyle \bigtriangleup}{=}\,\,}}
\newcommand{\<}{\left\langle}
\renewcommand{\>}{\right\rangle}
\newcommand{\herm}{{\sf H}}
\newcommand{\trasp}{{\sf T}}
\renewcommand{\vec}{{\rm vec}}
\newcommand{\transp}{{\sf T}}
\newcommand{\calL}{\mbox{${\mathcal L}$}}
\newcommand{\calO}{\mbox{${\mathcal O}$}}

\newcommand{\Afd}{\mbox{$\boldsymbol{\mathcal{A}}$}}
\newcommand{\Bfd}{\mbox{$\boldsymbol{\mathcal{B}}$}}
\newcommand{\Cfd}{\mbox{$\boldsymbol{\mathcal{C}}$}}
\newcommand{\Dfd}{\mbox{$\boldsymbol{\mathcal{D}}$}}
\newcommand{\Efd}{\mbox{$\boldsymbol{\mathcal{E}}$}}
\newcommand{\Ffd}{\mbox{$\boldsymbol{\mathcal{F}}$}}
\newcommand{\Gfd}{\mbox{$\boldsymbol{\mathcal{G}}$}}
\newcommand{\Hfd}{\mbox{$\boldsymbol{\mathcal{H}}$}}
\newcommand{\Ifd}{\mbox{$\boldsymbol{\mathcal{I}}$}}
\newcommand{\Jfd}{\mbox{$\boldsymbol{\mathcal{J}}$}}
\newcommand{\Kfd}{\mbox{$\boldsymbol{\mathcal{K}}$}}
\newcommand{\Lfd}{\mbox{$\boldsymbol{\mathcal{L}}$}}
\newcommand{\Mfd}{\mbox{$\boldsymbol{\mathcal{M}}$}}
\newcommand{\Nfd}{\mbox{$\boldsymbol{\mathcal{N}}$}}
\newcommand{\Ofd}{\mbox{$\boldsymbol{\mathcal{O}}$}}
\newcommand{\Pfd}{\mbox{$\boldsymbol{\mathcal{P}}$}}
\newcommand{\Qfd}{\mbox{$\boldsymbol{\mathcal{Q}}$}}
\newcommand{\Rfd}{\mbox{$\boldsymbol{\mathcal{R}}$}}
\newcommand{\Sfd}{\mbox{$\boldsymbol{\mathcal{S}}$}}
\newcommand{\Tfd}{\mbox{$\boldsymbol{\mathcal{T}}$}}
\newcommand{\Ufd}{\mbox{$\boldsymbol{\mathcal{U}}$}}
\newcommand{\Vfd}{\mbox{$\boldsymbol{\mathcal{V}}$}}
\newcommand{\Wfd}{\mbox{$\boldsymbol{\mathcal{W}}$}}
\newcommand{\Xfd}{\mbox{$\boldsymbol{\mathcal{X}}$}}
\newcommand{\Yfd}{\mbox{$\boldsymbol{\mathcal{Y}}$}}
\newcommand{\Zfd}{\mbox{$\boldsymbol{\mathcal{Z}}$}}